\documentclass{article}

\usepackage{algorithm}
\usepackage{algorithmic}
\usepackage{threeparttable}
\usepackage{amsmath,amssymb}
\usepackage{graphicx}
\usepackage{epstopdf}
\usepackage{PRIMEarxiv}
\usepackage{enumitem}
\usepackage[utf8]{inputenc} 
\usepackage[T1]{fontenc}    
\usepackage{hyperref}       
\usepackage{url}            
\usepackage{booktabs}       
\usepackage{amsfonts}       
\usepackage{nicefrac}       
\usepackage{microtype}      
\usepackage{lipsum}
\usepackage{fancyhdr}       
\usepackage{subcaption}      
\graphicspath{{media/}}     
\usepackage{cite}

\title{Act Better by Timing: A timing-Aware Reinforcement Learning for Autonomous Driving
}

\author{Guanzhou Li$^{1}$ \qquad Jianping Wu$^{1,2,3}$ \qquad Yujing He$^{1}$ \vspace{0.1em} \\
        $^1$Tsinghua University \quad $^2$Tsinghua University Research Institute at Shenzhen \quad $^3$Sichuan Tianfu Yongxing Laboratory \vspace{0.1em}\\
        {\tt \hspace{0mm}ligz19@mails.tsinghua.edu.cn  \quad \tt jianpingwu@tsinghua.edu.cn \quad \tt hyj19@mails.tsinghua.edu.cn}
}

\begin{document}
\maketitle

\begin{abstract}
    Autonomous vehicles inevitably encounter a vast array of scenarios in real-world environments. Addressing long-tail scenarios, particularly those involving intensive interactions with numerous traffic participants, remains one of the most significant challenges in achieving high-level autonomous driving. Reinforcement learning (RL) offers a promising solution for such scenarios and allows autonomous vehicles to continuously self-evolve during interactions. However, traditional RL often requires trial and error from scratch in new scenarios, resulting in inefficient exploration of unknown states. Integrating RL with planning-based methods can significantly accelerate the learning process. Additionally, conventional RL methods lack robust safety mechanisms, making agents prone to collisions in dynamic environments in pursuit of short-term rewards. Many existing safe RL methods depend on environment modeling to identify reliable safety boundaries for constraining agent behavior. However, explicit environmental models can fail to capture the complexity of dynamic environments comprehensively. Inspired by the observation that human drivers rarely take risks in uncertain situations, this study introduces the concept of action timing and proposes a timing-aware RL method, In this approach, a "timing imagination" process previews the execution results of the agent's strategies at different time scales. The optimal execution timing is then projected to each decision moment, generating a dynamic safety factor to constrain actions. A planning-based method serves as a conservative baseline strategy in uncertain states. In two representative interaction scenarios, an unsignalized intersection and a roundabout, the proposed model outperforms the benchmark models in driving safety.
\end{abstract}


\section{Introduction}

As an emerging technology, autonomous driving is expected to significantly improve traffic safety by preventing accidents caused by human errors such as distracted driving. Despite substantial research and investment, autonomous driving systems excel in many common scenarios but still fall short of human performance in continuous long-distance driving\cite{feng2023dense}, especially in long-tail scenarios involving intensive and unpredictable interactions.

The hand-crafted rules and deterministic algorithms, traditionally employed in the autonomous driving industry, struggle to handle the nearly infinite variety of road environments. This often results in overly cautious driving strategies that leave autonomous vehicles virtually immobilized in complex interactive situations. In contrast, learning-based methods offer an alternative approach, enabling the development of driving strategies through human imitation or direct interaction with the environment. The emerging end-to-end method unifying perception, planning, and decision-making, has demostrated adaptability across diverse driving conditions\cite{hu2023planning}. However, as the behavior of surrounding traffic participants also changes in response to the autonomous vehicle's decisions\cite{huang2023differentiable}, supervised learning requires extensive data to mimic accurate human behavior in interactive environments. Moreover, the rarity of data in risky scenarios can lead to autonomous vehicles failing to respond appropriately in safety-critical situations.

Reinforcement learning (RL), a method of self-evolution through environmental interaction, has achieved success in various domains\cite{silver2017mastering,kaufmann2023champion,cao2023continuous}, particularly in autonomous driving, where it can acquire practical interaction skills without requiring extensive labeled data. Beyond independent implementation, RL has proven effective in enhancing the robustness and reactive capabilities of pre-trained models developed through imitation learning or supervised learning approaches\cite{lu2023imitation}. As autonomous vehicles advance toward higher levels of intelligence, RL's role in refining their interactive capabilities becomes increasingly pivotal.

However, challenges remain in applying RL directly to highly dynamic environments like autonomous driving. In the RL framework, the agent (the ego vehicle) refines its strategy through continuous trial and error, aiming to maximize its rewards. Training the agent in interactive scenarios often relies on intricate reward design and faces a dilemma: low penalties for collisions might encourage reckless actions, while high penalties could result in overly cautious behavior that hinders task completion. The absence of intrinsic safety constraints can lead the agent into dangerous states, which is unacceptable in practical driving. To address this, safe reinforcement learning (Safe RL) incorporates a cost function to keep cumulative risks within acceptable limits\cite{achiam2017constrained,cao2021confidence,yang2021wcsac} or integrates a safety shield mechanism to check and mask unsafe actions\cite{ha2018recurrent,wang2022dynamic,wu2023daydreamer}. Yet, the unpredictable environmental dynamics will also pose risks. In the same state, surrounding traffic participants might select varied actions with different probabilities, causing shifts in the agent's expected optimal action trajectories. These trajectories often appear within limited time windows, and sometimes even slight deviations in actions can lead to undesired collisions. To capture environmental dynamics, the world model encodes the observations of RL into a low-dimensional latent space, enabling the agent to better understand and predict the environment and react accordingly.

World model-driven RL typically requires sufficient data and additional training to accommodate the environmental uncertainty, and the accuracy of the world model determines the effectiveness of the agent's actions. In densely interactive scenarios, possible occlusions and uncertain behavioral intensions of surrounding participants challenge explicitly modeling environmental dynamics. In such situations, human drivers often adopt more cautious strategy during uncertain, risky moments and act decisively during determined, safer moments. Inspired by this, we propose a timing-aware reinforcement learning (timing-aware RL) framework consisting of two agents, the "actor" and the "timing taker", along with a base planner. The actor agent develops its policy to pursue the highest far-sighted values, whereas the base planner responds to the dynamic environment with immediate, conservative strategies to mitigate potential conflicts at each moment. The timing taker evaluates the optimal timing for the actor's action by attempting to execute the action progressively over different time scales. Since the actor needs to select actions at each time step in response to the rapidly changing environment, the timing taker learns in a parallel environment where the actor's actions are sparsely executed, referred to as the "timing imagination". The timing determined by the timing taker will be projected to the next decision moment through an asymptotic function and provides a safe constraint on the actor's action. Simply put, the actor's action is preferred when its optimal timing is closer to the present moment, otherwise a more conservative and immediate baseline strategy is employed.

Timing-aware RL exploits the environmental dynamics to deeply explore the evolution of future states, which allows agents to discover high-value states more effectively in the case of sparse reward and to make the correct decisions by seizing the timing window in highly uncertain environments. Additionally, the proposed method allows the agent to search for the optimal policy in a space formed by both reward and timing dimension, which prevents the agent from becoming "lazy" or getting stuck in a local optimum due to the difficulty of discovering long-term optimal rewards in complex interactions. 

The main contributions of this study include:


\begin{itemize}
	\item[1] Proposing a timing-aware reinforcement learning framework that balances the pursuit of long-term value and the avoidance of immediate risk by combining learning-based and planning-based approaches through timing, enabling the agent to discover high-value actions by optimizing in reward and timing dimensions cyclically.
	\item[2] Introducing a "timing imagination" parallel to the real environment, where the timing taker learns the optimal timing by gradually executing the actor's actions at different time scales, and further providing safety constraints for actions in the real environment.
	\item[3] Validating the proposed method in two representative interactive scenarios, demonstrating better performance than advanced safe RL algorithms.
\end{itemize}

The rest of this paper is organized as follows: Section 2 introduces the application of safe RL in autonomous driving, Section 3 presents the problem statement for decision making in interactive driving scenarios, Section 4 details the specific methods of this study, Section 5 describes the experimental design and results, and Section 6 presents the conclusions and future prospects of this work.

\section{Related Works}
\subsection{RLs in interactive traffic scenarios}
Reinforcement learning is a method to achieve policy improvement in interaction with the environment. This learning paradigm is naturally suitable for application in scenarios where an autonomous vehicle interacts with other traffic participants, and is therefore widely used in autonomous driving tasks such as adaptive cruise control\cite{desjardins2011cooperative}, lane changing/overtaking\cite{mirchevska2018high}, and junction passage\cite{bautista2022autonomous}. However, in interactive traffic scenarios, RL methods are still challenged by uncertainty in the reactions of surrounding traffic participants and the diversity of interaction patterns. In order to better capture the possible responses of human drivers, game theory is introduced into the RL framework. For instance, Li et al. propose a Deep Q-Network considering Nash Equilibrium to develop the ramp merging strategy of an autonomous vehicle\cite{li2024nash}. In scenarios involving multiple vehicles, the Nash gaming model may not reach a stable equilibrium. Stackelberg gaming and level-k reasoning models can be applied to predict the intentions of the surrounding vehicles and help RL agent respond rationally\cite{yuan2021deep,rahmati2021helping,wang2022comprehensive}.

On the other hand, to meet the challenges of complex interaction patterns, some studies design elaborate architecture of neural networks to process complex observations in RL. Chen et al. introduce spatio-temporal attention into Deep Deterministic Policy Gradient (DDPG), enabling the agent to stay focused on interactions with the traffic participants that significantly influence the ego vehicle's motion\cite{chen2019attention}; Cai et al. employ Graph Attention Networks (GATs) to capture the relative relationships and interaction patterns between vehicles\cite{cai2022dq}. Besides, human demonstration can also guide RL agent to efficiently adapt to complex interactive scenarios. Huang et al. adopt Kullback-Leibler (KL) divergence to constrain the deviation between RL policies and expert policies\cite{huang2022efficient}; Kulkarni et al. propose hierarchical frameworks combining RL with imitation learning\cite{kulkarni2016hierarchical,le2018hierarchical,cao2020reinforcement}. Since human demonstration data is costly and not available in all scenarios, self-imitation allows the RL agent to mimic its past optimal decisions to develop robust strategies to form a robust baseline policy\cite{oh2018self}. In the absence of human demonstration data, the combination of planning-based methods is the other way to speed up the RL agent's learning process in a new scenario. Naveed et al. present a hierarchical RL framework: a Double DQN gives macro actions in the upper layer, and optimal trajectory is generated in the middle layer, then a PID controller keeps the autonomous vehicle driving along the trajectory in the lower layer\cite{naveed2021trajectory}.

In complex interaction scenarios, the distribution of positive rewards can be sparse and subject to variations due to environmental dynamics, such as unprotected left turns in dense traffic. This often leads agents to fall into local optima, becoming either overly aggressive, resulting in collisions, or overly conservative, causing excessive delays. Addressing such scenarios requires a clear understanding of the potential evolution of the environment. World model, as an emerging technology, empowers agents to understand and predict changes in the world. It encodes the observed variables into a low-dimensional hidden space via an autoencoder to eliminate environmental noises, enabling the agent to better extract core features in environmental dynamics and predict possible future states of the environment in the hidden space\cite{chen2019model,ma2021reinforcement,chen2021interpretable,wang2022dynamic}. Despite these advancements, predicting the dynamics of interactive environments remains challenging due to inherent uncertainties.

\subsection{RL for safe autonomous driving}

While many methods enhance the interactive capabilities of RL, most lack built-in mechanisms to ensure safety, which is paramount in autonomous driving. Safe RL has therefore gained significant attention, addressing the low tolerance for task failure in this domain. An intuitive way is to check the output of the RL agent and mask unsafe actions\cite{ma2021model}, or to project unsafe actions into safe sets\cite{cheng2019end,li2021safe}. Common practices include defining a Control Barrier Function (CBF) or Control Lyapunov Function (CLF) to realize smooth control of an autonomous vehicle and to avoid it falling into unsafe states\cite{zhang2023spatial}. Also, some rule-based methods, such as the finite state machine, can be integrated into RL to forbid unreasonable agent behavior\cite{nguyen2023safe}.

Risk-constrained RL formulates a Constrained Markov Decision Process(CMDP) and defines cost functions to quantify future risk levels, aiming to minimize risk or maintain cumulative risk below a threshold. For example, Li et al. employ a Bayesian risk assessment model that evaluates risk based on state uncertainties and relative distances between the ego vehicle and surrounding traffic, motivating the agent to learn a policy that minimizes expected risk\cite{li2022lane}. The Lagrangian method is commonly used to balance reward maximization and risk minimization via Lagrange multipliers\cite{schmidt2022learn}. However, policy oscillation issues in Lagrangian-based RL may degrade performance. To address this, dynamic adjustment of weight coefficients based on current and historical risk levels prevents overly conservative policies while maintaining safety\cite{peng2022model, gu2023safe}. An alternative approach, proposed in \cite{wang2023autonomous}, encourages actions close to the original strategy but below a predefined risk threshold. 

Robust RL strives to achieve robust policy enhancement under environmental uncertainty, such as Adversarial RL and Confidence-aware RL. Adversarial RL has demonstrated that introducing environmental noise or adversarial behaviors of other vehicles during agent training can effectively improve its ability to handle complex situations\cite{he2023fear, he2023towards}. Confidence-aware RL develops a baseline policy and activates learned policy only when it demonstrably outperforms the baseline policy with a specified confidence level \cite{cao2021confidence, zhou2022dynamically,cao2022trustworthy,cao2023continuous}. 

However, the aforementioned methods predominantly react passively to environmental risks and uncertainties, failing to accurately seize the timing of actions based on the evolving characteristics of the environment. This capability is indispensable in traffic interactions, as entering uncertain states at inappropriate times may prevent the agent from learning effective behavioral policies.

\section{Problem Statement}

\begin{figure}[htbp]
	\centering
	\includegraphics[width=0.45\textwidth]{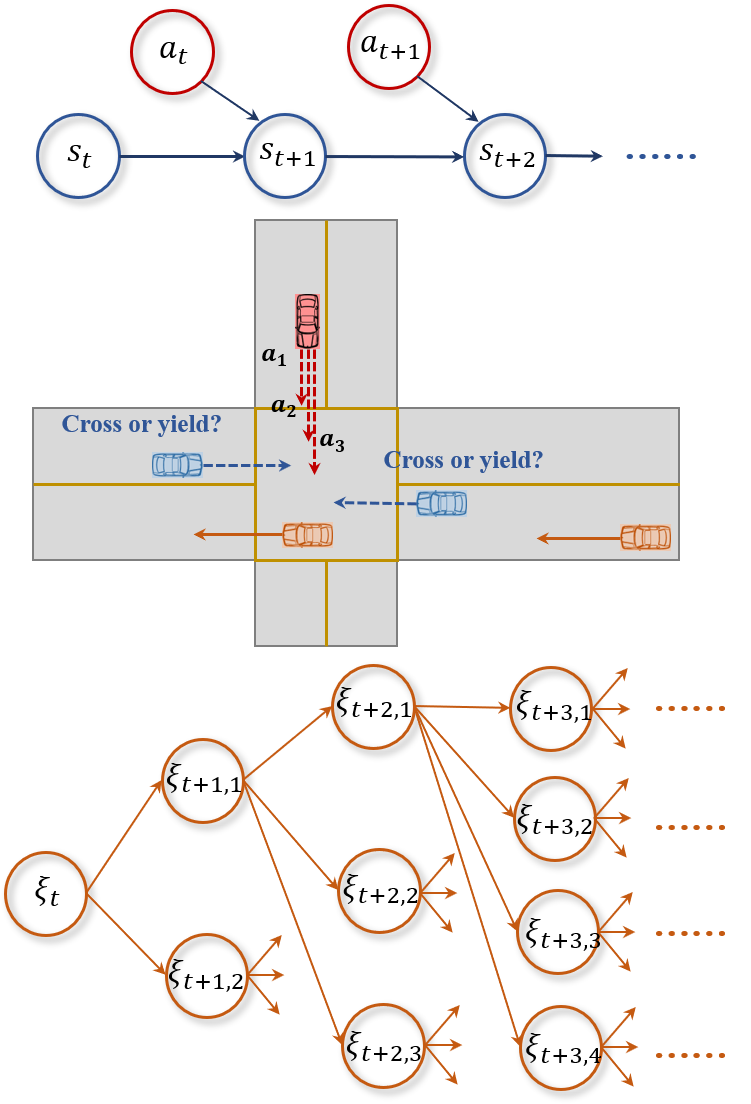}
	\caption{The interaction process between ego vehicle and the environment}
	\label{fig:fig1}
\end{figure}

The interaction of the autonomous vehicle with its surrounding can be represented as a Markov decision process (MDP) defined by five-variable tuple $(S, A, P, R, \gamma)$, where $S$, $A$, $P$, $R$ and $\gamma$ represent the state space, action space, transition probabilities between states, reward mechanism, and the discount factor for future rewards, respectively. In the context of dynamic interaction, the possible next state $s_{t+1}$ of the ego vehicle is jointly determined by the current state $s_t$, the selected action $a_t$, and the dynamics of the environment $\xi_t$, as expressed in Eq.\ref{eq:eq1}.

\begin{equation}
	\label{eq:eq1}
	s_{t+1} = f(s_t, a_t, \xi_t)
\end{equation}

where $\xi_t$ refers to the inherent dynamic nature and randomness of the environment, called the "environmental state" below. It is hardly affected by the ego vehicle's action in the current moment, like the two orange vehicles distant from the ego vehicle in Fig.\ref{fig:fig1}. According to Eq.\ref{eq:eq1}, $\xi_t$ is further specified as that part of the environmental dynamics associated with the state transitions of the ego vehicle throughout the driving process. Since the ego vehicle in different states will be affected by the environment dynamics to different degrees, the environmental state $\xi_t$ is also relevant to the current state $s_t$, as expressed in Eq.\ref{eq:eq2}.

\begin{equation}
	\label{eq:eq2}
	\xi_{t} = g(\xi_{t-1}, s_{t}, \epsilon_{t})
\end{equation}

where $\epsilon_{t}$ denotes the stochasticity at moment $t$. From Eq.\ref{eq:eq2}, the fact that the agent's action $a_t$ in regular RL does not directly act on environment state $\xi_t$, results in the effect of $\xi_t$ not being sufficiently considered when the agent learns it's strategy. However, the term $\xi_t$ will affect the rewards that the ego vehicle can ultimately receive and should not be ignored. For instance, the ego vehicle will have a better chance to cross the interweaving zone when traffic is sparse, while it faces more risks when traffic is dense. 

Since a stepwise decision process does not involve the optimization of environmental dynamics $\xi_t$, we introduce the concept of optimal execute timing $T^*$. For any given action $a_t$, this concept identify the most suitable opportunity to take it among multiple future environmental states $\left(\xi_{t+1}, \xi_{t+2}, \cdots, \xi_{t+T_{max}}\right)$. Based on this, we can construct a MDP with $T_{max}-1$ transferable states, as expressed in Eq.\ref{eq:eq3}, where the transfer probability between states can be expressed as Eq.\ref{eq:eq4}.

\begin{equation}
	\label{eq:eq3}
	\pi_{T}: \xi_t \rightarrow \left(\xi_{t+1}, \xi_{t+2}, \cdots, \xi_{t+T_{max}}\right)
\end{equation}

\begin{equation}
	\label{eq:eq4}
	P(\xi_{t+T}|\xi_t, T) = P(\xi_{t+T}|T, \xi_t, \xi'_{t+1}, \cdots, \xi'_{t+T-1})\cdots P(\xi'_{t+1}|T, \xi_t)
\end{equation}

Since $\xi_{t+1}$ is affected by $s_{t+1}$ and $s_{t+1}$ is related to $a_t$, the search for optimal timing requires decoupling decisions on action and timing. To realise this, a timing-dependent asymptotic function is introduced to represent the process by which $a_t$ is gradually executed. When considering both the evolution of environmental state $\xi_{t-1} \rightarrow \xi_t$ and the gradual execution of action as transfer probabilities inherent in the environment system, Eq.\ref{eq:eq3} can be formulated as a trainable form, as expressed in Eq.\ref{eq:eq5}.

\begin{equation}
	\label{eq:eq5}
	\pi_T: (s_t, a_t) \rightarrow ((s_{t+1}, a_{t+1}), \cdots, (s_{t+T_{max}}, a_{t+T_{max}}))
\end{equation}

In dynamic environments, by learning the laws of environmental evolution patterns described in Eq.\ref{eq:eq5}, the ego vehicle can reduce the impact of environmental uncertainty on decision making and better leverage the environment dynamics to improve the driving safety.


\section{Methodology}

\begin{figure}[htbp]
	\centering
	\includegraphics[width=\textwidth]{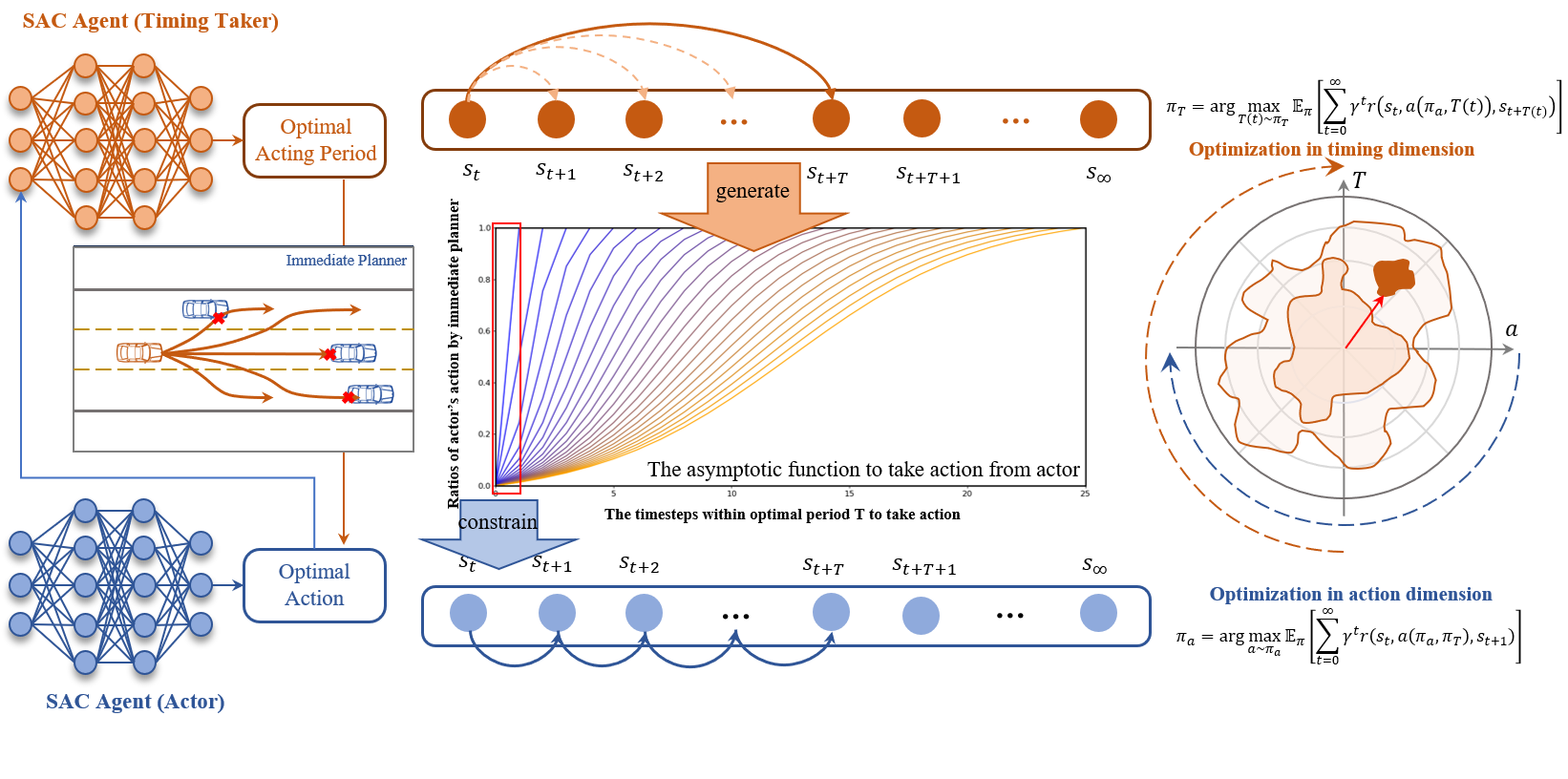}
	\caption{The Timing-aware Reinforcement Learning Framework}
	\label{fig:fig2}
\end{figure}

In this study, we propose a timing-aware reinforcement learning framework, where the optimal action for each step and the optimal time to take that action are solved by the "actor" and the "timing taker", respectively (as illustrated in Fig.\ref{fig:fig2}). Their strategies are denoted as $\pi_{\phi_a}$ and $\pi_{\phi_T}$, where the $\phi_a$ and $\phi_T$ denote the training parameters of the models. Since it is impossible for autonomous driving to be completely inactive in an dynamic environment, the ego vehicle adopts a sampling-based planning method as the baseline strategy and calculates the possible safety actions at each moment, denoted as $a_{base}\sim\pi_b$. As a "reactive" strategy, the baseline model simply makes a prediction and planning based on the current motion state of the surrounding objects, and it tends to be "short-sighted" and conservative. It can achieve real-time conflict avoidance, but it can also prevent vehicles from accomplishing their assigned tasks or cause collisions between other vehicles due to irrational interactions, so autonomous driving needs to balance the baseline strategy (considering short-term costs) with the actor's strategy (optimizing long-term returns). Aiming at this, the actual executed action of the ego vehicle is expressed as $\bar{a}=g(a_{actor}, T)$, where $T$ is the optimal execution timing of the actor's action $a_{actor}\sim\pi_{\phi_a}$. When $T$ is smaller, it means that the sooner execution of $a_{actor}$ can bring higher rewards, otherwise it means that the environment dynamics is not suitable for $a_{actor}$ currently, and the more conservative baseline action $a_{base}$ should be taken.

The training of timing-aware reinforcement learning is divided into three steps: the first step is to train an initial policy $\pi_{\phi'_a}$ of the “actor” without timing constraints, which is realized by the Soft Actor-Critic (SAC) model. The second step is to train the timing taker's policy $\pi_{\phi_T}$ based on $\pi_{\phi'_a}$, and the third step is to train the safe policy of the “ actor ” under the constraint of timing $T\sim\pi_{\phi_T}$. During the training process, the second and the third steps are carried out alternately. The training environment of the “ actor ” is consistent with the real environment, where actions are generated at each time step. In the training environment of the timing taker, the execution time of each action is treated as a decision step, and within each decision step, the ego vehicle's actions will gradually progress from the baseline strategy to the expected action. Since this process will not be ultimately executed in the real environment, the environment is called the "timing imagination".

In the timing imagination, the timing taker will learn the mapping from a given state and action to the optimal execution time of the action $\pi_{\phi_T}: (s_t, a_t)\rightarrow T_t\in[1, T_{max}]$, where $a_t$ represents the action selected by the actor at time $t$, and when $T=1$, it means that the action should be executed immediately at the next step. During the $T$ steps of the action execution period, the proportion $\beta$ of the ego-vehicle's strategy that employs $a_t$ gradually increases, and the proportion $1-\beta$ that employs the baseline strategy then gradually decreases. A hyperbolic tangent function is used to represent such a "timing-dependent asymptotic process" as shown in Eq.\ref{eq:eq6}, where $\omega$ is the shape coefficient that determines the steepness of the curve growth, and then the actual action of the ego vehicle at the moment $t+\Delta t$ can be expressed as Eq.\ref{eq:eq7}, when $\Delta t = T_t$, then $\beta(T_t, \Delta t)=1$, $\bar{a}_{t+\Delta t}=a_t$.

\begin{equation}
	\label{eq:eq6}
	\beta(T_t, \Delta t)=\frac{1}{2} \left(\frac{tanh(\omega(\frac{\Delta t}{T_t}-\frac{1}{2}))}{tanh(\frac{\omega}{2})} + 1\right), \Delta t = 1,2,\cdots, T_t
\end{equation}

\begin{equation}
	\label{eq:eq7}
	\bar{a}_{t+\Delta t} = \beta(T_t, \Delta t)\cdot a_t + (1-\beta(T_t, \Delta t))\cdot a_{base, t+\Delta t}
\end{equation}

The timing taker learns the MDP described in Eq.\ref{eq:eq5} through the SAC model, where the reward for each decision step is expressed as a sum of step rewards considering the discount factor, as shown in Eq.\ref{eq:eq8}. The Q Network is updated by Eq.\ref{eq:eq9}, where $\theta_T$ denotes the training parameters of the timing-taker's Q network, and $\hat{Q}_{\theta_T}$ is the target Q network according to the SAC framework, and the discount factor for each decision step is equal to $\gamma^{T_t}$.

\begin{equation}
	\label{eq:eq8}
	r_T(s_t, a_t, T_t) = \sum^{T_t}_{\Delta t=1} \gamma^{\Delta t-1} r(s_t,\bar{a}_{t+\Delta t})
\end{equation}

\begin{equation}
	\label{eq:eq9}
	Q_{\theta_T}(s_t,a_t,T_t)\leftarrow r_T(s_t, a_t, T_t)+\gamma^{T_t}\mathbb{E}_{s_{t+T}\sim P}[\hat{Q}_{\theta_T}(s_{t+T_t}, a_{t+T_t}, T_{t+T_t})]
\end{equation}


The above formulas keep the timing taker's assessment of the value of states in the timing imagination consistent with that of the actor in the real environment, thus enabling them to collaborate in finding the optimal solution in the dynamic interaction. The objective of the timing taker's policy network is to maximize the value of $Q_{\theta_T}$ by selecting the appropriae execution timing $T$ given the state-action pairs $(s,a)$, as shown in Eq.\ref{eq:eq10}. Following the implementation of SAC in \cite{haarnoja2018soft}, Eq.\ref{eq:eq10} contains an entropy term $\mathcal{H}(\pi_{\phi_T}(\cdot|s,a))=-\log \pi_{\phi_T}(T|s,a)$ and the temperature coefficient $\alpha$ is adaptively updated regarding the exploration degree of the environment.

\begin{equation}
	\label{eq:eq10}
	\max_{\phi_T} J(\phi_T)=\mathbb{E}_{(s,a)\sim\mathcal{D}, T\sim\pi_{\phi_T}}[Q_{\theta_T}(s,a,T)+\alpha\mathcal{H}(\pi_{\phi_T}(\cdot|s,a))]
\end{equation}

In the timing imagination, the attempt of the timing taker to act in the most appropriate environmental state $\xi_t$ can be viewed essentially as a delayed execution of $a_t$. However, since the action $a_t$ is determined in state $s_t$ at moment $t$, the delayed execution may suffer from the cumulative error due to the environmental stochasticity $\epsilon_t$, resulting in the action not necessarily being the optimal choice at step $t+\Delta t$. Therefore, the gradual execution of $a_t$ in the timing imagination is projected to each moment in the real environment to provide safety constraints for the actor. Specifically, $\Delta t$ in Eq.\ref{eq:eq6} is made equal to $1$ to get the scale factor, $\beta(T_t, 1)$, which indicates the suitability of the action $a_t$ in the current environmental state $\xi_t$. This factor is called the "timing factor" and abbreviated as $\beta(T_t)$ as in Eq.\ref{eq:eq11}, then the actual action adopted by the ego vehicle at each moment in the real environment is written as Eq.\ref{eq:eq12}.

\begin{equation}
	\label{eq:eq11}
	\beta(T_t) := \beta(T_t, 1) = \frac{1}{2} \left(\frac{tanh(\omega(\frac{1}{T_t}-\frac{1}{2}))}{tanh(\frac{\omega}{2})} + 1\right)
\end{equation}

\begin{equation}
	\label{eq:eq12}
	\bar{a}_{t} = \beta(T_t)\cdot a_t + (1-\beta(T_t))\cdot a_{base, t}
\end{equation}


The logic behind Eq.\ref{eq:eq12} is that the timing taker rehearses the execution trajectories of $a_t$ at different time scales in the timing imagination, and if $a_t$ is better suited for earlier execution, it means that $a_t$ has lower uncertainty and higher expected return in the current environmental state $\xi_t$, and thus $a_t$ should be chosen, otherwise the more conservative strategy $\pi_b$ is preferred. It is worth noting that although in Eq.\ref{eq:eq12}, $a_t\sim\pi_{\phi_a}$ and $a_{base, t}\sim\pi_b$ are combined by weighting, due to the application of hyperbolic tangent function, Eq.\ref{eq:eq12} can also be viewes as a "soft" choice between $a_t$ and $a_{base, t}$, which avoids the performance decay of the model that may result from a compromise strategy. When $T_t=1$, we have $\beta(T_t)=1$ and $\bar{a}_t=a_t$; As $T_t$ increases, $\beta(T_t)$ rapidly converges to zero, and $\bar{a}_t \approx a_{base,t}$; $\bar{a}_t$ takes the intermediate value between $\pi_{\phi_a}$ and $\pi_b$ only for the finite interval $T_t=2,3$. Besides, when $T_t\rightarrow T_{max}$, it does not require the actor to take action $a_t$ at the moment $T_{max}$, but rather it characterizes that the cautious action $a_{base, t}$ should be taken to wait for a better chance. Thus, projecting the process in the timing imagination into the "timing factor" allows the ego vehicle to act more flexibly without having to accept a suboptimal strategy due to the constraints of the predefined timing range $[1,T_{max}]$.

The training of the actor uses the SAC model constrained by the "timing factor". The actual action $\bar{a}_t$ at each step is given by Eq.\ref{eq:eq12}, and the corresponding reward is $r(s_t, \bar{a}_t)$. The Q network is updated as shown in Eq.\ref{eq:eq13}, where $\theta_a$ denotes its training parameters, and after several iterations, $\theta_a$ is copied to update the parameters of the target Q network $\hat{Q}_{\theta^-_a}$. The optimization objective of the actor's policy network is expressed as Eq.\ref{eq:eq14}, and $\phi_a$ is the training parameters.

\begin{equation}
	\label{eq:eq13}
	\max_{\theta_a} J(\theta_a) = \mathbb{E}_{s_{t+1}\sim P}\left[\left(r(s_t, \bar{a}_t)+\gamma \max_{a_{t+1}}\hat{Q}_{\theta^-_a}(s_{t+1},a_{t+1})-Q_{\theta_a}(s_t, a_t)\right)^2\right]
\end{equation}

\begin{equation}
	\label{eq:eq14}
	\max_{\phi_a} J(\phi_a)=\mathbb{E}_{s_t\sim\mathcal{D},a_t\sim\pi_{\phi_a}}\left[Q_{\theta_a}(s_t, a_t)+\alpha\mathcal{H}(\pi_{\phi_a}(\cdot|s_t))\right]
\end{equation}

The overall training process of timing-aware RL can be described as Alg.\ref{alg:alg1}

\begin{algorithm}
	\caption{Training of Timing-aware Reinforcement Learning}
	\label{alg:alg1}
	\begin{algorithmic}[1]
		\STATE Initialize parameters $\phi_a, \theta_a, \phi_T, \theta_T$
		\STATE The first step is train a regular SAC model as actor $\phi_a\leftarrow \phi'_a,\theta_a\leftarrow \theta'_a$
		\WHILE{iteration $<$ max training steps}
			\FOR{substep from $0$ to $N_T$}
				\STATE get action from the actor $a_t\sim\pi_{\phi_a}(a_t|s_t)$
				\STATE get timing from the timing taker $T_t\sim\pi_{\phi_T}(T_t|s_t, a_t)$
				\FOR{$\Delta t$ from $0$ to $T_t$}
					\STATE take asymptotic action $\bar{a}_{t+\Delta t}$ in timing imagination according to Eq.\ref{eq:eq7}
					\STATE get the next state $s_{t+\Delta t + 1}$ and reward $r(s_{t+\Delta t}, \bar{a}_{t+\Delta t})$
				\ENDFOR
				\STATE calculate reward $r_T(s_t, a_t, T_t)$ according to Eq.\ref{eq:eq8}
				\STATE push sample into memory: $\mathcal{D}_T\leftarrow\mathcal{D}_T \cup\{((s_t, a_t), T_t, r_T(s_t, a_t, T_t), s_{t+T_t}, \delta)\}$
				\STATE update the Q network of the timing taker: $\theta_T\leftarrow\theta_T-\lambda_{\theta_T}\hat{\nabla}_{\theta_T}J(\theta_T)$
				\STATE update the policy network of the timing taker: $\phi_T \leftarrow \phi_T-\lambda_{\phi_T}\hat{\nabla}_{\phi_T}J(\phi_T)$
			\ENDFOR
			
			\FOR{substep from $0$ to $N_a$}
				\STATE get action from the actor $a_t\sim\pi_{\phi_a}(a_t|s_t)$
				\STATE get timing from the timing taker $T_t\sim\pi_{\phi_T}(T_t|s_t, a_t)$
				\STATE take action $\bar{a}_{t}$ according to Eq.\ref{eq:eq12}
				\STATE get next state $s_{t+1}$ and reward $r(s_t, \bar{a}_{t})$
				\STATE push sample into memory: $\mathcal{D}_a\leftarrow\mathcal{D}_a \cup\{(s_t, a_t, r(s_t, \bar{a}_{t}), s_{t+1}, \delta)\}$
				\STATE update the Q network of the actor: $\theta_a\leftarrow\theta_a-\lambda_{\theta_a}\hat{\nabla}_{\theta_a}J(\theta_a)$
				\STATE update the policy network of the actor: $\phi_a \leftarrow \phi_a-\lambda_{\phi_a}\hat{\nabla}_{\phi_a}J(\phi_a)$
			\ENDFOR
		\ENDWHILE
	\end{algorithmic}
\end{algorithm}

In Alg.\ref{alg:alg1}, the actor's policy $\pi_{\phi_a}$ and the timing taker's policy $\pi_{\phi_T}$ are optimized iteratively. After the above training process, timing-aware RL can obtain a strategy superior to regular RL, expressed as Eq.\ref{eq:eq15}, where $\pi_{\phi'_a}$ denotes the policy of regular RL.

\begin{equation}
	\label{eq:eq15}
	\mathbb{E}_{\pi^*_{\phi_a}}\left(\sum^{\infty}_{t=0}\gamma^t r(s_t, \bar{a}_t, \xi_t)\right) \geq \mathbb{E}_{\pi_{\phi'_a}}\left(\sum^{\infty}_{t=0}\gamma^t r(s_t, a_t, \xi_t)\right)
\end{equation}

The proof of Eq.\ref{eq:eq15} requires to introduce assumption that, in timing-aware RL, the expected cumulative reward for the action constrained by "timing factor" (i.e., the $\bar{a}_t$ calculated by Eq.\ref{eq:eq12}), is greater than that of the action which is decided at the historical step $t-\Delta t$ and executed at the current moment $t$ according to Eq.\ref{eq:eq7} (abbreviated as $\bar{a}_{t-\Delta t}(t)$), as shown in Eq.\ref{eq:eq16}.
\begin{equation}
	\label{eq:eq16}
	\mathbb{E}_{\pi^*_{\phi_a}}\left(\sum^{\infty}_{t=0} \gamma^t r(s_t,\bar{a}_t,\xi_t)\right) \geq \mathbb{E}_{\pi^*_{\phi_a}}\left(\sum^{\infty}_{t=0} \gamma^t r(s_t,\bar{a}_{t-\Delta t}(t),\xi_t)\right)
\end{equation}

Eq.\ref{eq:eq16} holds because $\bar{a}_t$ in the left term the optimal choice of the actor considering the current state, while $a_{t-\Delta t}(t)$ may lose its optimality in the most recent state due to the uncertainty of the environmental evolution $(s_{t-\Delta t},\xi_{t-\Delta t})\rightarrow (s_t, \xi_t)$. Thus, the proof of Eq.\ref{eq:eq15} is given as follows.

\begin{equation}
\label{eq:eq17}
\begin{aligned}
	& \mathbb{E}_{\pi_{\phi_a}^*}\left(\sum_{t=0}^{\infty} \gamma^t r\left(s_t, \bar{a}_t, \xi_t\right)\right) \\
	& =\mathbb{E}_{\pi_{\phi_a}^*}\left[r\left(s_0, \bar{a}_0, \xi_0\right)+\gamma r\left(s_1, \bar{a}_1, \xi_1\right)+\cdots+r^k r\left(s_k, \bar{a}_{k^{\prime}}, \xi_k\right)+\cdots\right] \\
	& =\mathbb{E}_{\pi_{\phi_a}^*}\left[\left(r\left(s_0, \bar{a}_0, \xi_0\right)+\cdots+\gamma^T r\left(s_{T^*}, \bar{a}_{T^*}, \xi_{T^*}\right)\right)+\sum_{t=T^*+1}^{\infty} \gamma^t r\left(s_t, \bar{a}_t, \xi_t\right)\right] \\
	& \geq \mathbb{E}_{\pi_{\phi_a}^*}\left[\left(r\left(s_0, \bar{a}_0, \xi_0\right)+\cdots+\gamma^T r\left(s_T, \bar{a}_0\left(T^*\right), \xi_T\right)\right)+\sum_{t=T^*+1}^{\infty} \gamma^t r\left(s_t, \bar{a}_t, \xi_t\right)\right] \\
	& \geq \mathbb{E}_{\pi_{\phi_a}^{\prime *}}\left[r\left(s_0, a_0, \xi_0\right)+\sum_{t=1}^{\infty} \gamma^t r\left(s_t, a_t, \xi_t\right)\right] \\
	& =\mathbb{E}_{\pi_{\phi_a}^{\prime *}}\left(\sum_{t=0}^{\infty} r\left(s_t, a_t, \xi_t\right)\right)
\end{aligned}
\end{equation}

In Eq.\ref{eq:eq17}, the first inequality applies Eq.\ref{eq:eq16}, and the second ineuality expresses the optimization goal of the timing taker in the timing imagination. For action $a_0$, the timing taker will choose the optimal execution period $T^*$, and the execution of $T^*$-step trajectory $(s_0,\xi_0)\rightarrow\bar{a}_0\rightarrow\cdots\rightarrow\bar{a}_{T^*}\rightarrow(s_{T^*},\xi_{T^*})$, are expected to obtain higher reward than the direct execution of $a_0$ at the next moment.

The base model consists of a finite state machine (FSM) and a lattice planner. The FSM decides whether to “drive”, “slow down to stop” or “ brake in emergency” according to the results of path planning and the distance from the front vehicle. When a suitable driving path cannot be found, the planner first searches whether there is a valid path for slowing down to stop, and if there is still no available path, the vehicle will brake urgently. Given a reference line along the road, the lattice planner represents the trajectory of the vehicle with quintic polynomials in respect to time, and decouples the lateral and longitudinal displacements in frenet coordinates relative to the reference line, respectively. Taking the longitudinal displacement as an example, the longitudinal displacement of the ego vehicle relative to the reference line within the next $t$ timesteps can be written as Eq.\ref{eq:eq18}

\begin{equation}
	\label{eq:eq18}
	l(t)=c_1 t^5 + c_2 t^4 + c_3 t^3 + c_4 t^2 + c_5 t + c_6
\end{equation}

Where the unknown coefficients $c_1, c_2, \cdots, c_6$ can be solved based on the ego vehicle's current motion state and the desired state at the end of the trajectory. The constraint expression is shown in Eq.\ref{eq:eq19}.

\begin{equation}
	\label{eq:eq19}
	\left\{\begin{array}{l}
		l(0)=c_6=l_0 \\
		\dot{l}(0)=c_5=v_{l 0} \\
		\ddot{l}(0)=2 c_4=a_{l 0} \\
		l\left(t_e\right)=c_1 t_e^5+c_2 t_e^4+c_3 t_e^3+c_4 t_e^2+c_5 t_e+c_6=l_e \\
		\dot{l}\left(t_e\right)=5 c_1 t_e^4+4 c_2 t_e^3+3 c_3 t_e^2+2 c_4 t_e+c_5=v_{l e} \\
		\ddot{l}\left(t_e\right)=20 c_1 t_e^3+12 c_2 t_e^2+6 c_3 t_e+2 c_4=a_{l e}
	\end{array}\right.
\end{equation}

Where $\dot{l}(\cdot)$, $\ddot{l}(\cdot)$ represent the first-order and second-order derivatives to time. $(l_0, v_{l0}, a_{l0})$ denote the current position, velocity, and acceleration of the ego vehicle along the longitudinal direction (here called $l$-axis). Correspondingly,  $(l_e,v_{le},a_{le})$ represent the vehicle's motion state at the end of the trajectory along the $l$-axis, and $t_e$ is the planned trajectory duration. The terminal acceleration $a_{le}$ is set to $0$, indicating that the vehicle reaches a stable state at the end of trajectory. By sampling the three variables  $t_e, l_e, v_e$, the six unknown coefficients in Eq.\ref{eq:eq17} can be determined through the six equations in Eq.\ref{eq:eq18}, uniquely defining a candidate trajectory. Similarly, the lateral displacement trajectory can be derived for the vehicle. Among the various candidate trajectories generated by sampling different values of $t_e$, $l_e$ ($d_e$) and $v_{le}$ ($v_{de}$) in both longitudinal and lateral directions, the cost function for each trajectory is calculated using Eq.\ref{eq:eq19}. The trajectory with the lowest cost is selected, as shown in Eq.\ref{eq:eq20}, provided it satisfies kinematic and dynamic constraints.

\begin{equation}
	\label{eq:eq20}
	\min c(t_e, l_e, v_{le}, d_e, v_{de}) = K_1 \varphi_{coll} + K_2 \varphi_v + K_3 \varphi_t + K_4 \varphi_d + K_5 \varphi_{jerk}
\end{equation}

Where $K1$ to $K5$ are the weighting coefficients, and $\varphi_{coll}$, $\varphi_v$, $\varphi_t$, $\varphi_d$, and $\varphi_{jerk}$ denote the cost terms of the candidate trajectories in term of potential collision, speed expectation, planning duration, lateral offset, and maneuverability, as given in Eq.\ref{eq:eq21}.

\begin{equation}
	\label{eq:eq21}
	\begin{array}{l}
		\varphi_{coll} = \delta_{traj}\\
		\varphi_{v} = \left|\sqrt{v^2_{le} + v^2_{de}}-\hat{v}\right|\\
		\varphi_{t} = t_e\\
		\varphi_{d} = |d_e - d_0| + |d_e - d_{ref}|\\
		\varphi_{jerk} = \sum^{t_e}_{t=0}(|l'''(t)| + |d'''(t)|) 
	\end{array}
\end{equation}

Where $\delta_{traj}$ is taken as $1$ when there is conflicts between the planned trajectory of the ego vehicle and the predicted trajectories of the surrounding vehicles, and $0$ otherwise; $\hat{v}$ is the target speed of the ego vehicle, which is a predefined value for the "driving" state and $0$ for the "decelerate to stop" state; $\varphi_{t}$ is the term of planning duration, used to enhance the planning speed and the driving agility; $\varphi_{d}$ denotes the lateral offset of the end of the trajectory relative to current position of the ego vehicle $d_0$ as well as the reference line $d_{ref}$; and $\varphi_{jerk}$ represents the vehicle's accumulated braking operations, which are used to control the smoothness of the trajectory and improve the driving comfort; $l'''(t)$ and $d'''(t)$ denote the third order derivatives of the planned trajectory to the lateral and longitudinal coordinates. Through the above planning process, the acceleration at the next step $a_{t1}$ is constrained by the Intelligent Driver Model (IDM) and then used as the base action $a_{base}$, as shown in Eqs.\ref{eq:eq22}-\ref{eq:eq23}.

\begin{equation}
	\label{eq:eq22}
	a_{t1} = \sqrt{\ddot{l}^2(t_1) + \ddot{d}^2(t_1)}
\end{equation}

\begin{equation}
	\label{eq:eq23}
	a_{base} = \min(a_{t1}, a_{\text{IDM}})
\end{equation}

\section{Experiment and Results}

\subsection{Driving Scenarios}
\begin{figure}[htbp]
	\centering
	\includegraphics[width=0.8\textwidth]{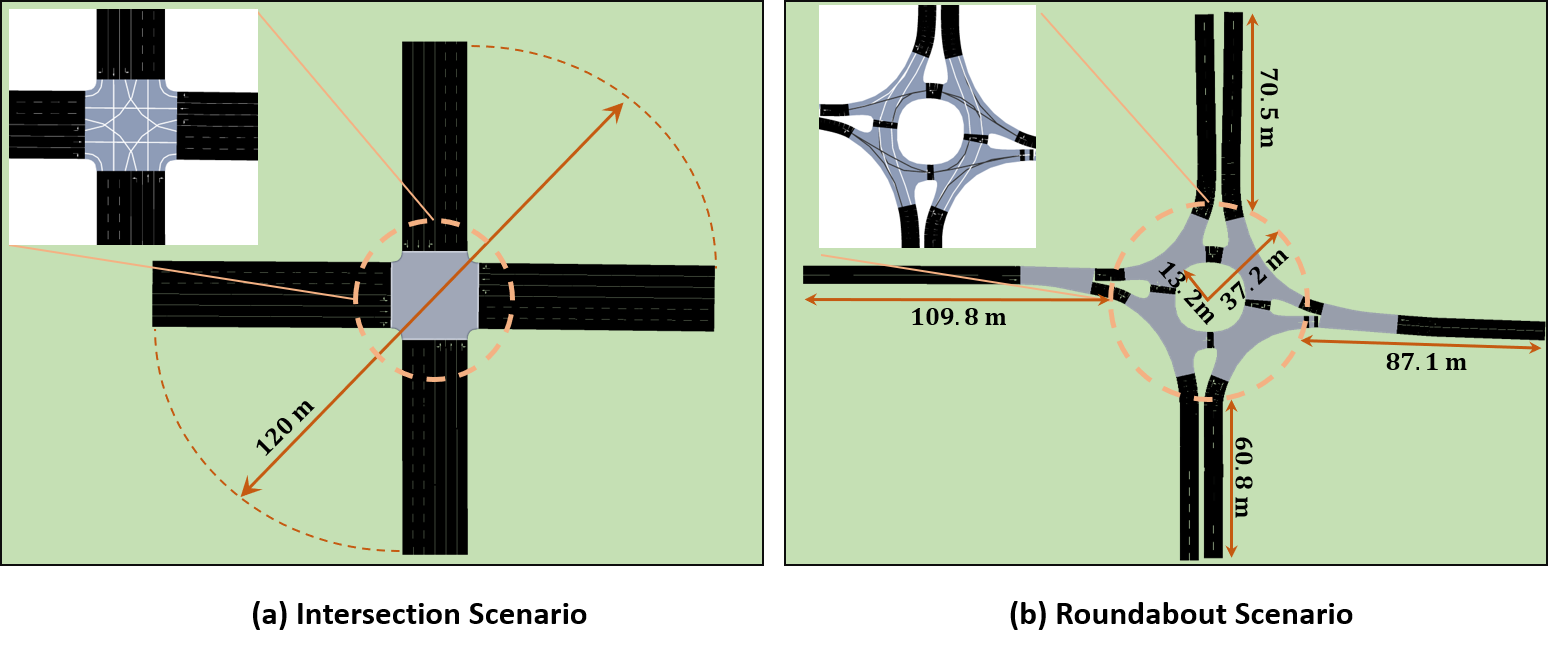}
	\caption{Road Networks for Interactive Experiments}
	\label{fig:fig3}
\end{figure}

Two most typical interaction scenarios for autonomous driving are selected to validated the proposed model: an unsignalized intersection and a roundabout, as shown in Fig\ref{fig:fig3}. In the intersection scenario, the entry road in each direction contains three lanes: right-turn, straight-only, and left-turn. Vehicles are spawned in any lane at the start of the road, which is $60\text{m}$ from the center of the intersection. In the roundabout scenario, the entry roads from the north and south directions have two lanes, and those from the east and west directions have only a single lane. The exact dimensions of the roundabout are labeled in Fig\ref{fig:fig3}(b). The simulation of traffic flow is realized by widely-used simulator SUMO\cite{lopez2018microscopic}, where the Intelligent Driver Model (IDM) is selected to simulate the car following process on the regular roads, and the parameters of IDM for each vehicle are randomly generated from a specified range, as shown in Tab.\ref{tab:tab1}. And we apply a level-k two-player game to model the interaction of human-driven vehicles with the ego vehicle during their approach to and passage through the conflict region. The utility function of game model takes safety, efficiency, comfort, and the patience of driver into account, and can be expressed as Eq.\ref{eq:eq24}.

\begin{equation}
	\label{eq:eq24}
	U = \alpha_1\varphi_{\text{safe}} + \alpha_2\varphi_{\text{eff}} + \alpha_3\varphi_{\text{comf}} + \alpha_4\varphi_{\text{emo}}
\end{equation}

Where $\alpha_1, \alpha_2, \alpha_3, \alpha_4$ are the coefficients of the utility function; $\varphi_{\text{safe}}$ denotes the time difference between the arrival of the vehicle and its conflicting vehicle at the conflict point; $\varphi_{\text{eff}}$ indicates the time the vehicle waiting for its conflicting vehicle to pass through the conflict point if it decides to yield, otherwise this term takes $0$ if it decides to cross; $\varphi_{\text{emo}}$ is computed from the waiting time of the vehicle before it enters the stop line or passes through the conflict point. All the above terms are normalized by Eq.\ref{eq:eq25}.

\begin{equation}
	\label{eq:eq25}
	\varphi_{i} \leftarrow \frac{(\varphi_i-\varphi_{i, \text{min}})}{(\varphi_{i,\text{max}}-\varphi_{i, \text{min}})}
\end{equation}

Level-k reasoning is a practical method to stabilizing game outcomes, where players make decisions alternately, with the k-level decision based on the opponent's $(k-1)$-level choice. As presented in \cite{wang2022comprehensive}, the value of k can typically set to $2$ for most traffic scenarios, a conclusion adopted here. The Human-driven vehicle first makes a level-0 choice--cross or yield--without considering its opponent (i.e. the autonomous vehicle). It assume the autonomous vehicle then makes a level-1 decision based on the utility function informed by the level-0 outcome. Finally, the human-driven vehicle evaluates the level-1 decision and reach a level-2 outcome. At level-2, the human-driven vehicle's selfish and altrustic attributes are measured by social value orientation (SVO) and considered into the utility function, as shown in Eq.\ref{eq:eq25.1}, where $U_{self}$ and $U_{oppo}$ denote the utility functions of the human-driven vehicle and the autonomous vehicle, respectively.

\begin{equation}
	\label{eq:eq25.1}
	U_{svo} = (1-\alpha_{svo}) U_{self} + \alpha_{svo} U_{oppo}
\end{equation}

In the experiment, to stabilize the traffic flow, the simulation will continuously run for a period of time $t_{sim}$, and the generation of traffic flow is controlled by the expected spawn time $t_{spawn}$. The spawn time describes the time interval between a newly departed vehicle and the previous one in the road network, generated from a probability distribution with expectation $t_{spawn}$. In each round of training or testing, the autonomous vehicle is randomly spawned in an available lane and reset when reaching the destination of its path.

\begin{table}[htb]
	\centering
	\begin{tabular}{llll}
		\toprule[1.5pt]
		\makebox[0.15\textwidth][l]{Scenario}  & \makebox[0.1\textwidth][l]{Parameters}   & \makebox[0.1\textwidth][l]{Values}  & \makebox[0.4\textwidth][l]{Explanations}\\ 
		\midrule[1.0pt]
		Both          & $L_{car}$     & $5.0 \text{m}$    & Vehicle length\\
		& $W_{car}$   & $2.0 \text{m}$  & Vehicle width\\
		& $W_{lane}$  & $3.2 \text{m}$  & Lane width \\
		& $a_{\text{max}}$  & $1.5\sim3.0 \text{m}/\text{s}^2$  & Maximum acceleration of the vehicle\\
		& $b_{\text{min}}$  & $-4.5\sim-2.0\text{m}/\text{s}^2$ & Maximum deceleration of the vehicle\\
		& $\hat{d}_{\text{min}}$    & $2.0\sim 4.0 \text{m}$   & Minimum gap distance from the front vehicle\\
		& $\hat{v}_{cf}$    &$8.0\sim12.0 \text{m}/\text{s}$  & Target speed of the car following process\\
		& $\alpha_1$ & $0.5\sim0.8$   & Safe Coefficient of the gaming model\\
		& $\alpha_2$ & $0.05\sim0.2$  & Efficient Coefficient of the gaming model\\
		& $\alpha_3$ & $0.05\sim0.2$  & Comfort Coefficient of the gaming model\\
		& $\alpha_4$ & $0.05\sim0.2$  & Emotion Coefficient of the gaming model\\
		& $\alpha_{svo}$ & $0.0\sim0.5$  & Social Value Oritentation of the vehicle\\
		& $k$ & $2$  & Highest decision level in the level-k game\\\midrule[1.0pt]
		Intersection  &$\hat{v}_{cr}$  &$4.5\sim6.0\text{m}/\text{s}$  & Target speed when crossing the intersection\\
		& $t_{sim}$ & $5000\text{s}$  & Time to reset the simulation\\
		& $t_{spawn}$ & $1.8\text{s}$  & Expected spawn gap of vehicles\\\midrule[1.0pt]
		Roundabout & $t_{sim}$ & $1000\text{s}$  & Time to reset the simulation\\
		&$t_{spawn}$  &$1.2\text{s}$  & Expected spawn gap of vehicles\\
		\bottomrule[1.5pt]
	\end{tabular}
	\caption{Environment Parameters}
	\label{tab:tab1}
\end{table}

\subsection{Configurations of the Autonomous Vehicle}
\begin{figure}[htbp]
	\centering
	\includegraphics[width=0.4\textwidth]{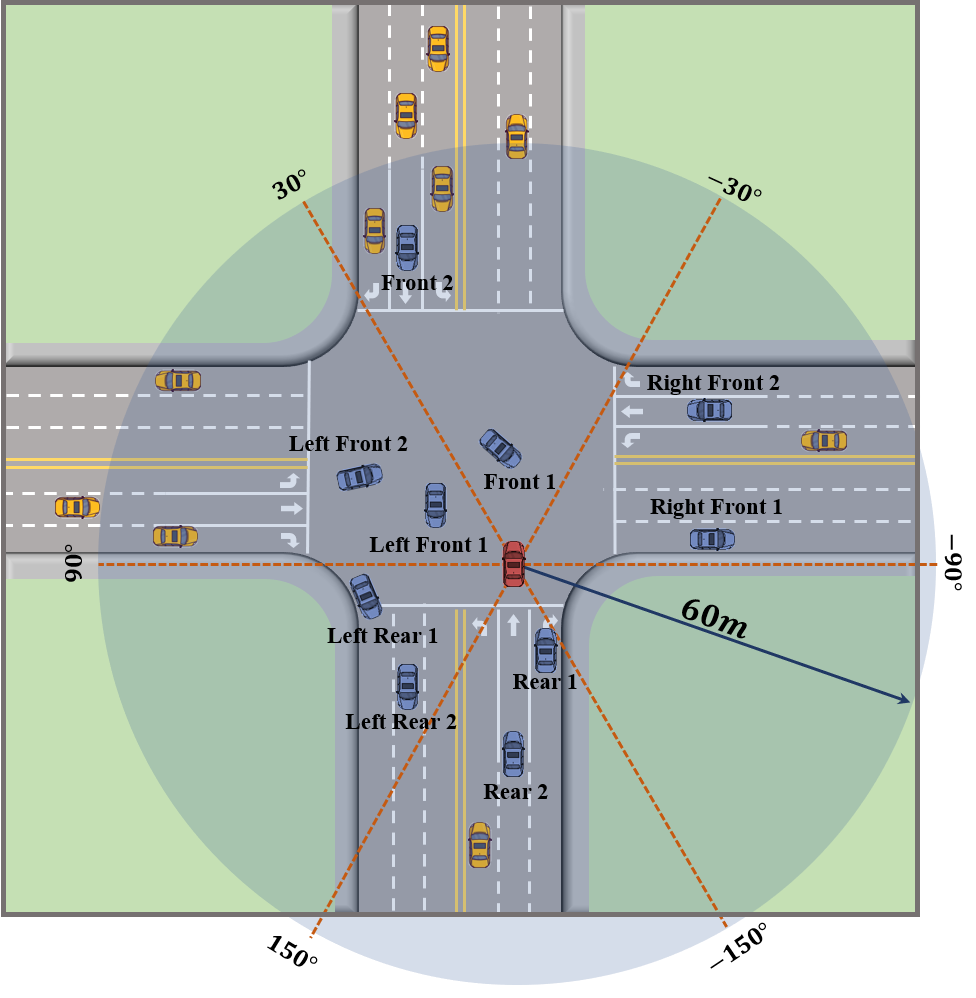}
	\caption{Observations of the autonomous vehicle}
	\label{fig:fig4}
\end{figure}

As illustrated in Fig.\ref{fig:fig4}, the observation of the ego vehicle has a radius of $60\text{m}$ and is divided into six areas according to the angular range of the vehicle's position relative to the ego vehicle: front$(-30^{\circ},30^{\circ}]$, left front$(30^{\circ}, 90^{\circ}]$, right front$(-90^{\circ}, -30^{\circ}]$, left rear$(90^{\circ}, 150^{\circ}]$, right rear$(-150^{\circ}, -90^{\circ}]$, and rear$(-180^{\circ}, -150^{\circ}]\cup (150^{\circ}, 180^{\circ}]$. In each area, the motion states of the two vehicles closest to the ego vehicle are included in the observation: the standardized distance to the ego vehicle $\hat{d}$, the standardized speed $\hat{v}$, the standardized position angle $\hat{\varphi}$, and the standardized heading angle $\hat{\theta}$, which can be expressed as Eq.\ref{eq:eq26}.

\begin{equation}
	\label{eq:eq26}
	\begin{array}{l}
		\hat{d} =\frac{d}{d_{max}}\\
		\hat{v} =\frac{v}{v_{max}}\\
		\hat{\varphi} =\frac{(\varphi-\varphi_{low})}{(\varphi_{high} -\varphi_{low})}\\
		\hat{\theta} =\frac{\theta -\theta_{ego}+\pi}{2\pi}
	\end{array}
\end{equation}

Where $d_{max}$ is taken as the perception range of the ego vehicle; $v_{max}$ denotes the maximum speed of the vehicle; $\varphi_{high}, \varphi_{low}$ are the upper and lower bounds of the position angle of each divided observation area, and $\theta, \theta_{ego}$ are the heading angles of observed vehicle and the ego vehicle. The ego vehicle's observation on its surroundings contains twelve 5-tuples.

\begin{equation}
	\label{eq:eq27}
	s_{\text{sur}}=\{s_{\text{sur}, ij}\} = \{(\delta_{ij},\hat{d}_{ij}, \hat{v}_{ij}, \hat{\varphi}_{ij}, \hat{\theta}_{ij})\}, i=1,2,\cdots,6; j=1,2
\end{equation}

In Eq.\ref{eq:eq26}, $s_{\text{sur}, ij}$ denote the motion state of the observed vehicle in the $i$-th observation area, and $j$ is ordered by the distance to the ego vehicle; $\delta_{ij}$ indicates the presence of vehicle in the $i$-th area, and takes $1$ if it does, oherwise it equals to $0$. In addition to the observations on the surrounding vehicles, the motion state of the ego vehicle should also be included in the training of RL, expressed as Eq.\ref{eq:eq28}. The input state of the agent is concatenated from $s_{\text{sur}}$ and $s_{\text{sur}}$, as given in Eq.\ref{eq:eq29}.

\begin{equation}
	\label{eq:eq28}
	s_{\text{ego}} = (h_1, h_2, \hat{v}_{ego})
\end{equation}

\begin{equation}
	\label{eq:eq29}
	s = (s_{\text{ego}}, s_{\text{sur}})
\end{equation}

In our experiment, the ego vehicle travels along one of the preset trajectories and only the speed term $\hat{v}_{ego}$ is included in the state $s_{\text{ego}}$, and $\hat{v}_{ego}$ can be calculated in the same way as Eq.\ref{eq:eq26}. $h_1, h_2$ are two one-hot vectors, and $h_1$ indicates the driving task of the ego vehicle, e.g. driving straight, turning left, and turning right; $h_2$ indicates the driving position, e.g., on the entry road, in the weaving area, on the exit road. 

In each episode of RL, given the driving path is fixed, the agent manipulates the motion of the ego vehicle via acceleration, which is limited by the maximum acceleration $2\text{m}/\text{s}^2$ and deceleration $-3\text{m}/\text{s}^2$. The agent will receive a final positive reward when the ego vehicle reaches the destination of the path, or it will be punished when the ego vehicle collides with surrounding vehicls. To improve the efficiency under normal driving conditions, the ego vehicle is incentivized to drive within a desired speed range $[v_{\text{min}}, v_{\text{max}}]$, thus the step reward of the environment can be expressed as Eq.\ref{eq:eq30}.

\begin{equation}
	\label{eq:eq30}
	r (s, a) = 
	\begin{cases}
		20,   &\text{if the ego vehicle reaches the destination}\\
		-20,  &\text{if the ego vehicle collides}\\
		\min(\frac{v_{ego}-v_{\text{min}}}{v_{\text{max}}-v_{\text{min}}}, 1)\times 0.5,   &\text{otherwise}
	\end{cases}
\end{equation}

In RL, the training of the agent lasts for $5\times10^5$ steps, and the agent's policy and Q networks both use the simple MLP architecture. The details of training parameters are given in Tab.\ref{tab:tab2}.

\begin{table}[htb]
	\centering
	\begin{tabular}{lll}
		\toprule[1.5pt]
		\makebox[0.1\textwidth][l]{Parameters}   & \makebox[0.1\textwidth][l]{Values}  & \makebox[0.4\textwidth][l]{Explanations}\\ 
		\midrule[1.0pt]
		$N_{train}$     &    $5\times10^5$     &  Total training steps in RL\\
		$B$             &    $256$             &  Batch size of the neural networks\\
		$N_{buffer}$    &    $1\times10^6$     &  Size of replay buffer in off-policy RL\\
		$N_{start}$     &    $2\times10^3$     &  Sampling steps before training the agent\\
		$D_{hidden}$    &    $256$             &  Dimensions of hidden layers in MLP\\
		$\gamma$        &    $0.99$            &  Discount factor of the future rewards\\
		$\tau$          &    $0.005$           &  Weight coefficient of the soft updating in SAC\\
		optimizer       &    Adam              &  Optimizer of the RL's networks\\
		lr              &    $3\times10^{-4}$  &  Learning rate of the RL's networks\\
		\bottomrule[1.5pt]
	\end{tabular}
	\caption{Training Parameters}
	\label{tab:tab2}
\end{table}

\subsection{Baseline Models}

To benchmark timing-aware RL, we implement four effective safe RL methods for comparison:
\begin{itemize}
	\item[1] CPO (Constrained Policy Optimization) is a policy optimization algorithm that ensures constraints are satisfied by directly incorporating them into the policy update considering the trust region.
	\item[2] PPO-Lag is developed upon Proximal Policy Optimization (PPO) and introduces a Lagrange relaxation to handle constraints, dynamically adjusting the penalty multiplier to balance reward maximization and constraint satisfaction.
	\item[3] SAC-Lag combines the Lagrange relaxiation with the highly competive off-policy RL algorithm SAC, which can realize a steady increase in rewards earned by the agent without violating constraints under a wide range of tasks.
	\item[4] SAC-PID is an extension of SAC-Lag, incorporating a Proportional-Integral-Derivative (PID) controller to dynamically tune the Lagrange multiplier, improving the stability of constraint satisfactation in the environments with varying safety requirements.
\end{itemize}

Moreover, in order to demonstrate the effectivenss and necessity of the timing-aware method, two variants of the proposed model are also tested in the experiments:

\begin{itemize}
	\item[1] Lattice-IDM, integrating the sampling-based planner and IDM model, is used as the baseline strategy $\pi_b$ in the proposed method.
	\item[2] SAC-Lattice utilizes the SAC model and additionally provides the action generated from the baseline strategy $\pi_b$ as optional action for the agent in each decision step.
\end{itemize}

\subsection{Results and Discussions}


\begin{figure}[htbp]
	\centering
	\begin{subfigure}{\textwidth}
		\centering
		\includegraphics[width=\textwidth]{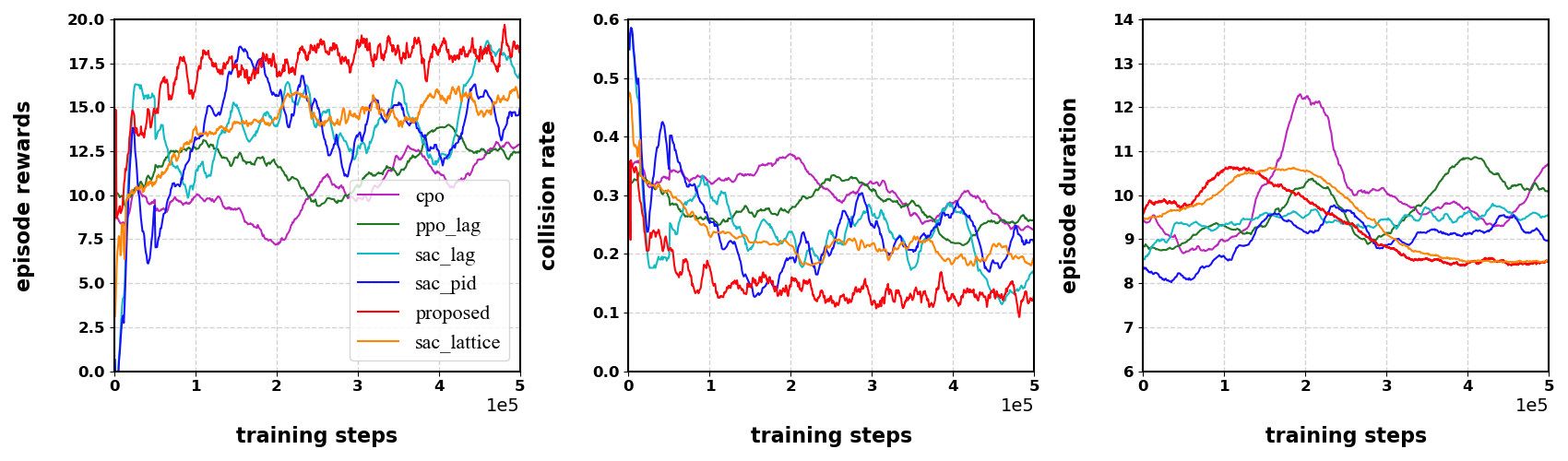}
		\caption{Intersection Scenario}
	\end{subfigure}
	\begin{subfigure}{\textwidth}
		\centering
		\includegraphics[width=\textwidth]{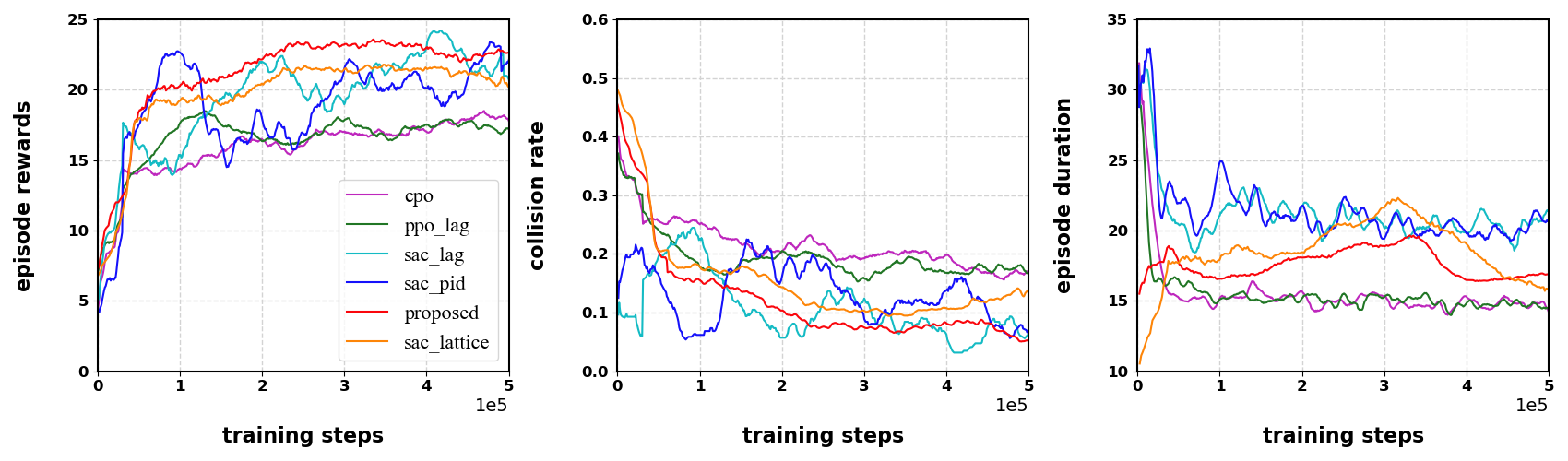}
		\caption{Roundabout Scenario}
	\end{subfigure}
	\caption{Training Curves}
	\label{fig:fig5}
\end{figure}

Fig.\ref{fig:fig5} illustrates the training performance of the proposed model with the six learning-based methods in the two experimental scenarios. In both scenarios, the proposed model gains the highest episode rewards and achieves the lowest collision rate compared to the competetive baseline methods. Meanwhile, since the planning module provides a relatively robust optional strategy for the agent's learning process, both the proposed model and SAC-lattice converge rapidly, and in particularly, the proposed model can obtain stably high returns after $1\times10^5$ training steps in both scenarios. In successive simulations, as traffic flow flucatates, queues of waiting vehicles are more likely to occur at an unsignalized intersection than at a roundabout. Under the conditions of varying traffic flow densities, the agent faces different difficulties in crossing the weaving area, and thus the performance curves in scenario 1 have more apparent fluctuations than those in scenario 2 in Fig.\ref{fig:fig5}. The combination of planning-based and learning-based methods can effectively reduce the instability of performance during the learning process caused by environmental changes. In terms of efficiency, the proposed model takes the shortest time to pass the complex unsignalized intersection scenario, and takes a similar amount of time as the fastest-passing model in the roundabout scenario, but with a lower collision rate. When training is completed, for each scenario, we conduct ten rounds of evaluation experiments for all models separatively, and each round of evaluation experiment includes $200$ test cases, to compute the mean and standard deviation in success rate and crossing time of each model, the results are shown in Tab.\ref{tab:tab3}.

\begin{table}[!htb]
	\centering
	\begin{tabular}{lcc|cc}
		\toprule[1.5pt]
		\makebox[0.1\textwidth][l]{} & \multicolumn{2}{c|}{\makebox[0.3\textwidth][c]{Intersection}}   & \multicolumn{2}{c}{\makebox[0.3\textwidth][c]{Roundabout}}\\
		Models & Success Rate & Crossing Time  & Success Rate & Crossing Time\\\midrule[1.0pt]
		CPO      & $0.769\pm0.038$ & $11.901\pm0.335$ & $0.838\pm0.029$ &$15.437\pm0.238$\\
		PPO-Lag  & $0.739\pm0.039$ & $10.303\pm0.515$ & $0.797\pm0.016$ &$15.366\pm0.181$\\
		SAC-Lag  & $0.828\pm0.026$ & $9.230\pm0.134$ & $0.872\pm0.040$ &$19.185\pm0.334$\\
		SAC-PID  & $0.810\pm0.028$ & $9.208\pm0.133$ & $0.907\pm0.016$ &$19.324\pm0.109$\\\midrule[1.0pt]
		Proposed & $0.881\pm0.021$ & $8.473\pm0.145$ & $0.912\pm0.025$ &$16.066\pm0.095$\\
		SAC-Lattice & $0.808\pm0.032$ & $8.509\pm0.182$ &$0.868\pm0.037$ &$16.371\pm0.481$\\
		Lattice-IDM & $0.896\pm0.018$ & $11.349\pm0.280$ & $0.797\pm0.030$ &$27.293\pm0.291$\\
		\bottomrule[1.5pt]
	\end{tabular}
	\caption{Results of Comparative Experiments}
	\label{tab:tab3}
\end{table}

In performance evaluations, the proposed model outperforms all learning-based baselines. In the intersection scenario, it achieve an $88.1\%$ task success rate, improving by $14.6\%$, $19.2\%$, $6.4\%$, and $8.8\%$ relative to CPO, PPO-Lag, SAC-Lag, and SAC-PID, respectively. Ablation study shows that the integrating a timing-aware mechanism enhance safety metric by $9.0\%$ compared to a simple combination of learning-based and planning-based method (i.e., SAC-Lattice). Although in this scenario, the proposed model exhibits a slight reduction in safety ($-1.8\%$) compared to its baseline safety strategy which serves as a fallback, it significantly improves the efficiency by $25.4\%$, striking a practical balance between safety and task efficiency. Interestingly, while the planning-based baseline is designed as the safety alternative, the safety performace of the proposed model is not strictly constrained by the baseline's limitations. In the roundabout scenario, the increased road curvature, reduced navigable space, and more merge-in-and-out points complicates trajectory prediction and poses more challenges in planning task, which makes the safety performance of the planning-based method (i.e. Lattice-IDM) drop noticably. Nevertheless, the proposed model achieves the highest success rate at $91.2\%$, surpassing all learning-based models with improvements of $8.8\%$, $14.4\%$, $4.6\%$, $0.5\%$, and $5.1\%$ over CPO, PPO-Lag, SAC-Lag, SAC-PID, and SAC-Lattice, respectively. These results demonstrate that selecting appropriate actions at the right time maximizes the complementary strengths of learning- and planning-based methods. Moreover, the integration of RL effectively addresses the "action freezing" problem in the planning-based method, which denotes the prolonged stops when no candidate paths is feasible caused by the aforementioned challenges. The hybrid strategy of planning and learning can effectively overcome this drawback, and the proposed model reduces crossing time by $41.1\%$ compared to the purely planning-based method.

\subsection{Case Analysis}

\begin{figure}[htbp]
	\centering
	\begin{subfigure}{\textwidth}
		\centering
		\includegraphics[width=\textwidth]{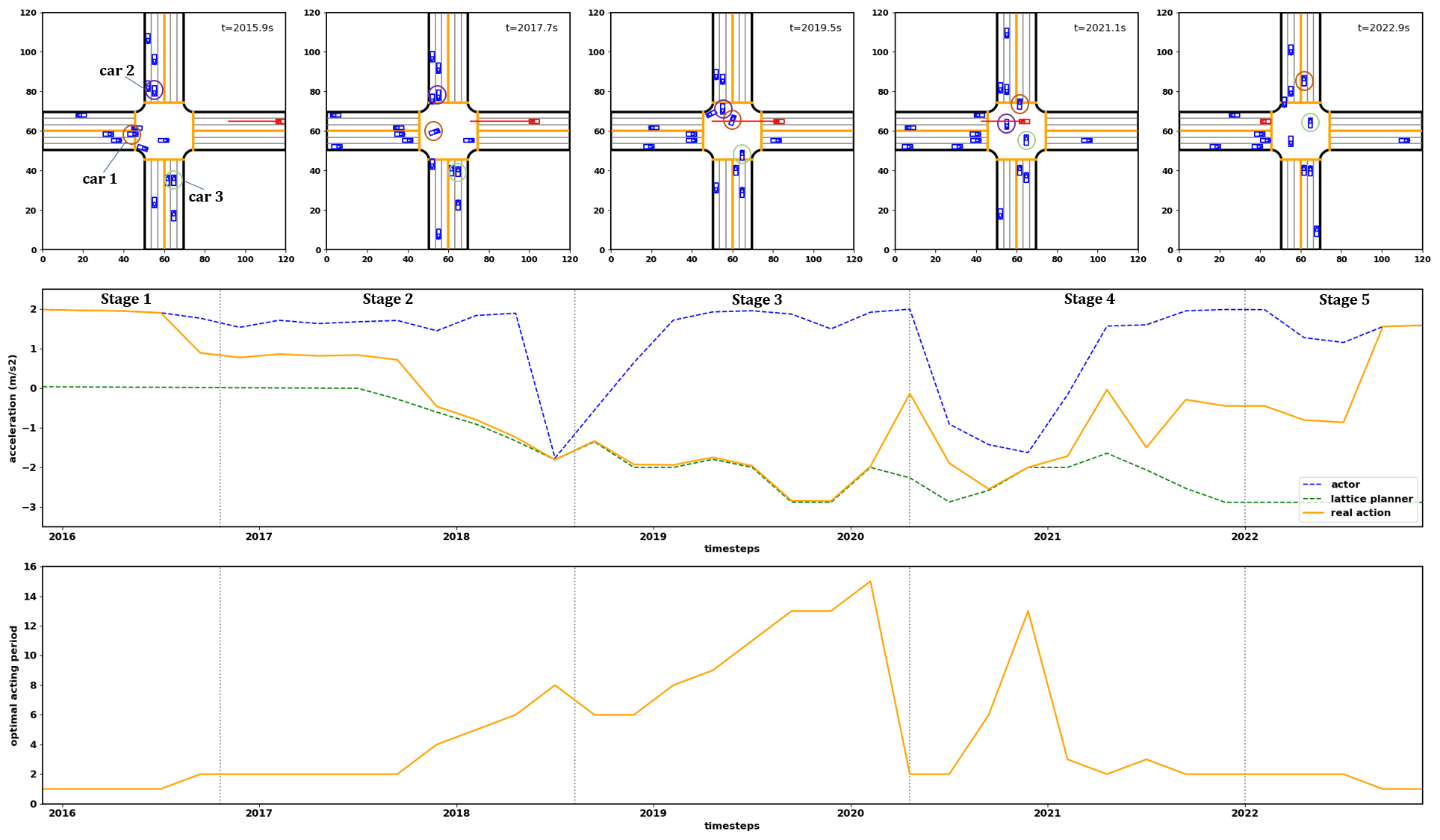}
		\caption{Intersection Scenario}
	\end{subfigure}
	\begin{subfigure}{\textwidth}
		\centering
		\includegraphics[width=\textwidth]{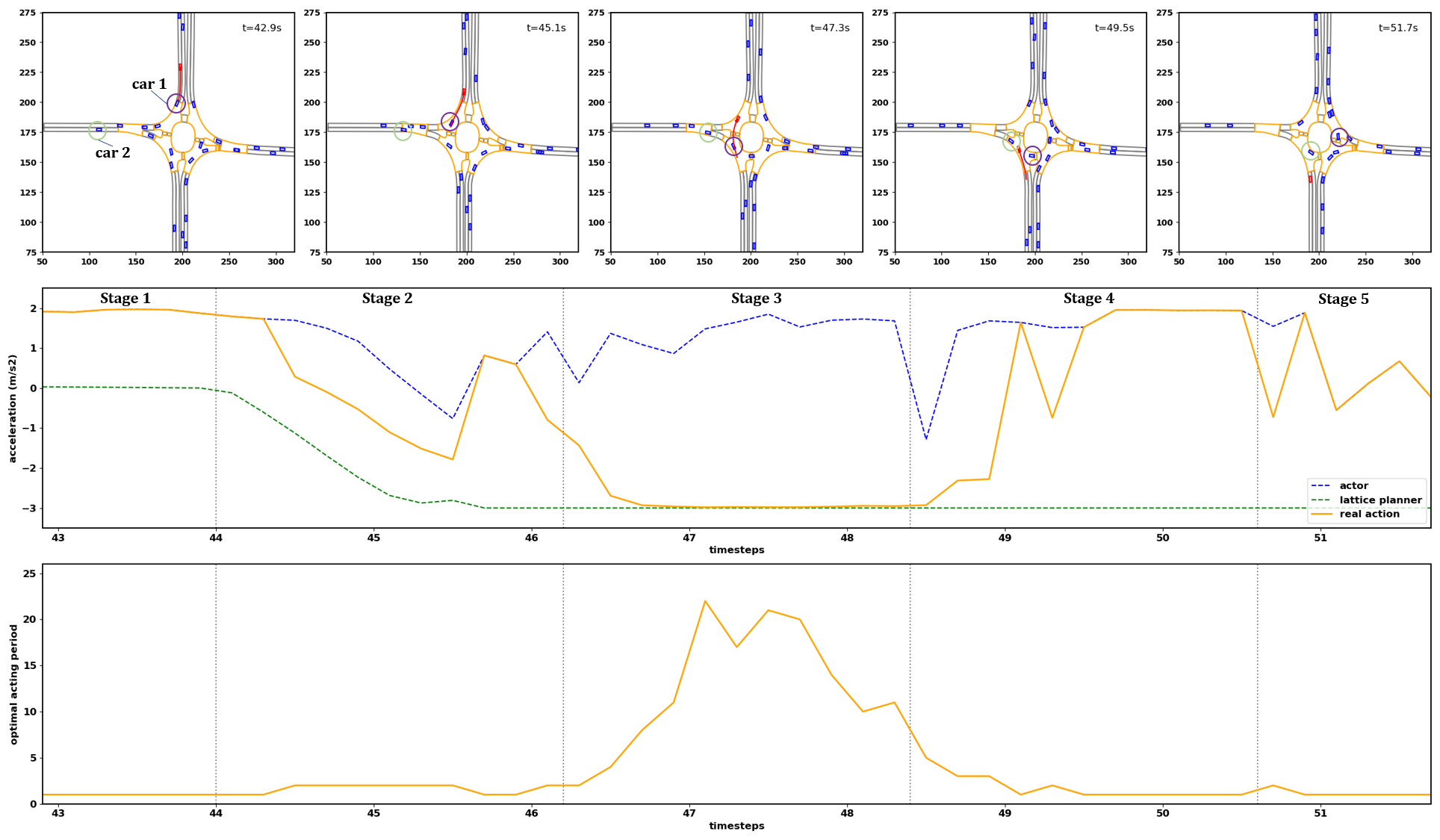}
		\caption{Roundabout Scenario}
	\end{subfigure}
	\caption{Driving Cases in the experiment}
	\label{fig:fig6}
\end{figure}

To visually illustrate the agent's driving behavior in two driving scenarios, we selected one test case each for the intersection and roundabout scenarios, as shown in Fig.\ref{fig:fig6}. In each scenario, the top five subfigures present snapshots of the autonomous vehicle at five equally spaced moments in the process from departure to exiting the conflict area. The autonomous vehicle is labeled in red, and the red trajectory is the planning result of the baseline planner. The middle subfigure exhibits the actions given by the actor (blue dashed line) and the baseline planner (green dashed line) at each moment, as well as the synthesized actions actually executed by the autonomous vehicle as calculated by Eq.\ref{eq:eq12} (orange solid line). The bottom subfigure gives the timing-taker's judgement of the optimal timescale to execute the action generated by the actor at each moment.

In the intersection scenario shown in Fig.\ref{fig:fig6}(a), the ego vehicle initally accelerates as suggested by the actor in Stage 1, as there are no other vehicles around it. From the Stages 2 to 3, as the ego vehicle approaches the conflict area, it adopts the cautious action given by the base planner, decelerating to observe and wait for Car 1 to cross first. By the end of Stage 3, the ego vehicle accelerates to claim the conflict point ahead of Car 3, then decelerates to allow Car 2 to pass before exiting the intersection. From this process, it can be seen that timing-aware RL naturally generate fine-grained actions to adapt to the inter-vehicle game in dense traffic, which can be challenging for planning-based methods (e.g., in scenarios like unprotected left turns, planning-based method might encounter excessive waiting problem). 

In the roundabout scenario illustrated in Fig.\ref{fig:fig6}(b), the ego vehicle starts with acceleration to follows Car 1. Upon reaching the weaving area of the roundabout in Stage 2, it decelerates slightly to keep a safe distance from the preceding Car 1 and to observe the surrounding environment for any potential conflicts. In Stage 3, the ego vehicle approaches the conflict point with Car 2, and the two vehicles are at a similar distances to the conflict point, which is prone to collisions, so the agent decides to yield. When Car 2, impeded by the halted Car 1 ahead, also stops, the gap between Cars 1 and 2 allows the ego vehicle to proceed. As a result, in Stage 4, the agent makes several exploratory accelerations and then quickly passing the conflict point. In this scenario, it can be observed that the actor prefers a more aggressive interaction strategy, while the strategy of base planner appears to be overly conservative and inflexible in the face of the complicated environmental dynamics (see the action curve of the base planner stays braked and barely fluctuates). Timing-aware RL can effectively balance learning- and planning-based methods to achieve a safer and more flexible interaction between the ego vehicle and the surrounding vehicles.

\begin{figure}[htbp]
	\centering
	\begin{subfigure}{\textwidth}
		\centering
		\includegraphics[width=\textwidth]{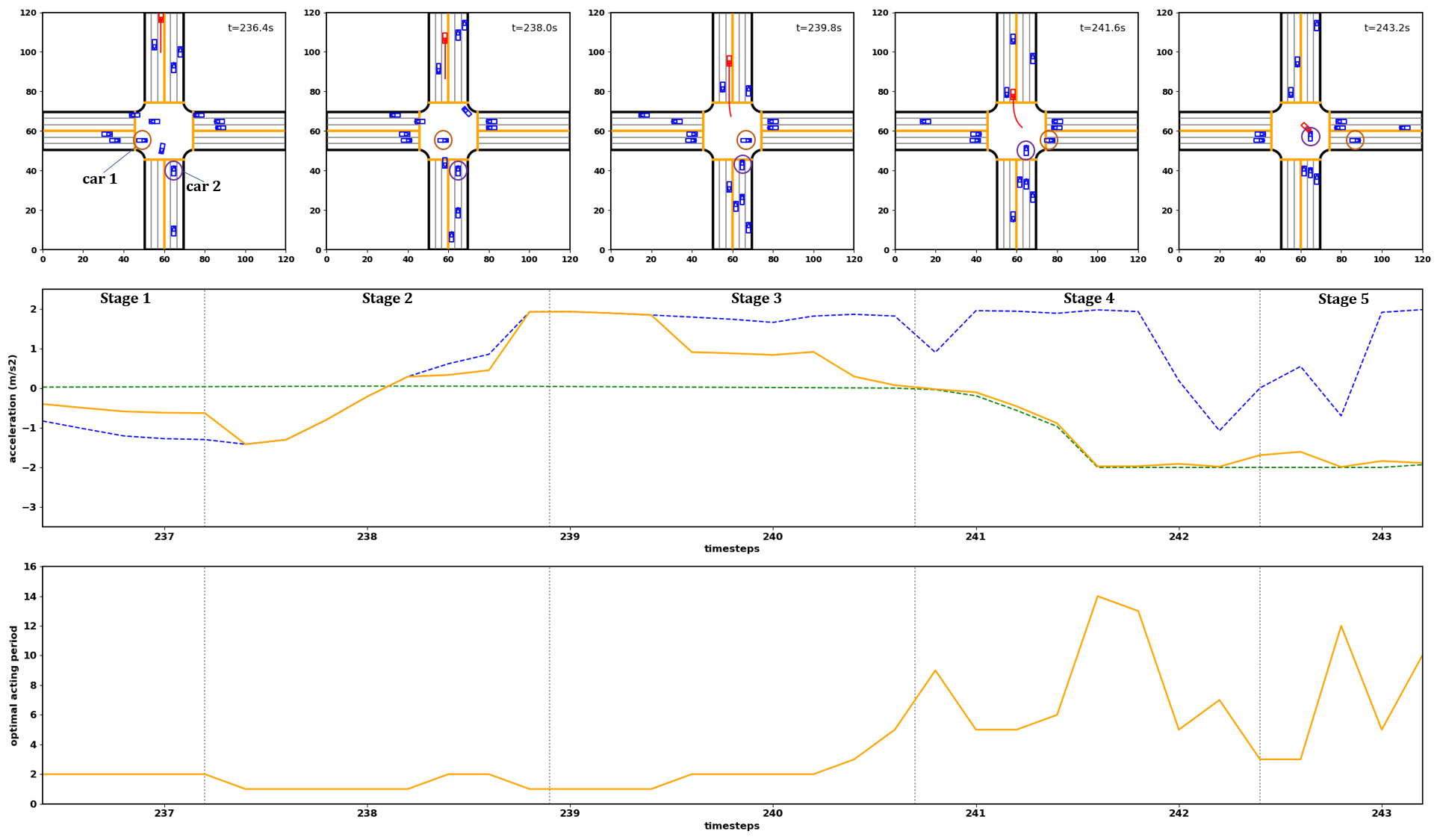}
		\caption{Intersection Scenario}
	\end{subfigure}
	\begin{subfigure}{\textwidth}
		\centering
		\includegraphics[width=\textwidth]{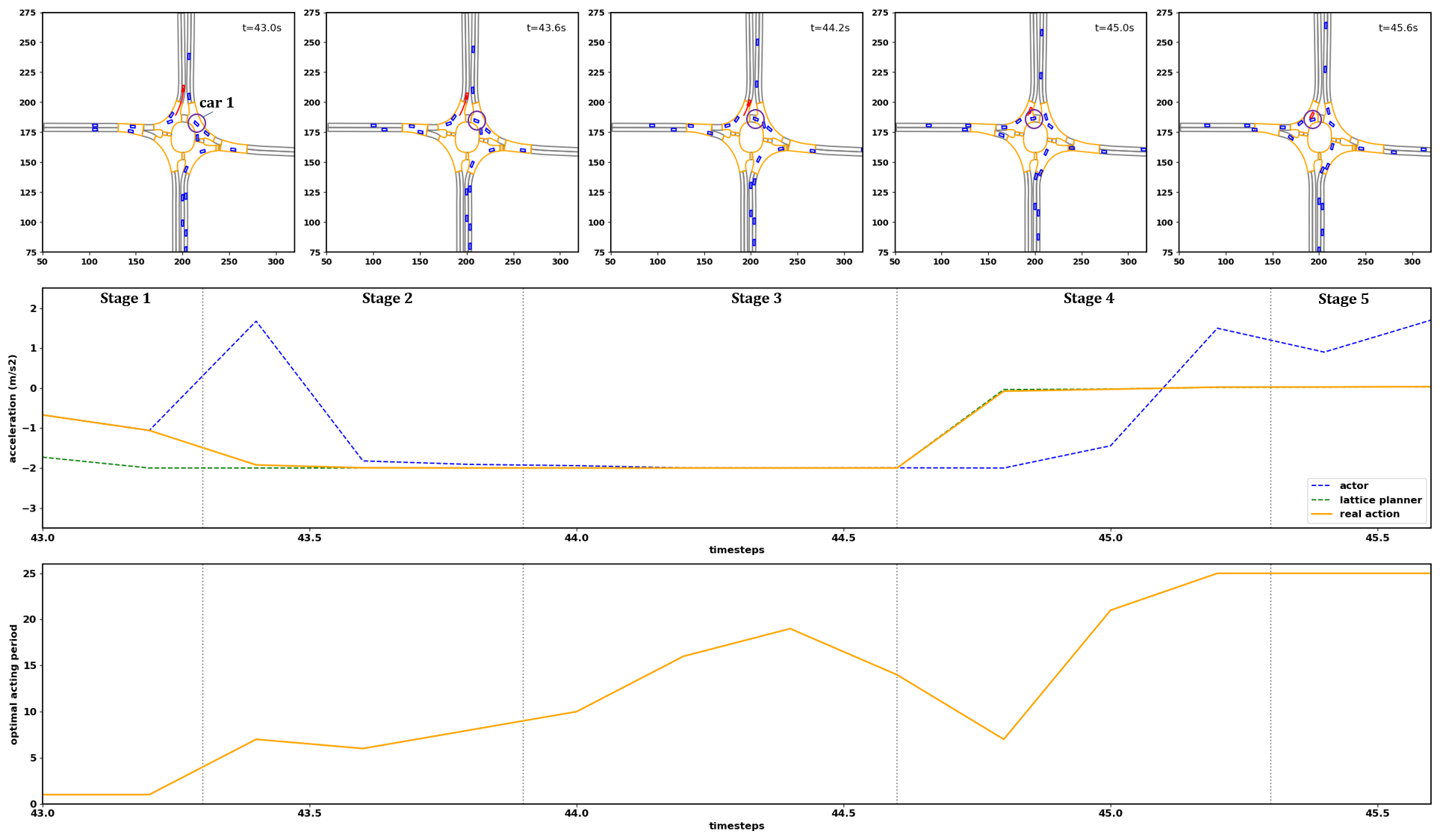}
		\caption{Roundabout Scenario}
	\end{subfigure}
	\caption{Failure Cases in the experiment}
	\label{fig:fig7}
\end{figure}

To better clarify the reasons for model failure and to provide subsequent improvement that can be made, we have also analyzed the failure cases, as shown in Fig.\ref{fig:fig7} for the intersection and roundabout scenarios, respectively. In the first scenario, before Car 1 exits the intersection, Car 2 waits at the stop line. Since the base planner assumes the surrounding vehicles move at a constant speed, and its planned trajectory does not conflict with the stopped Car 2, thus it suggests the ego vehicle to maintain a fixed speed from Stages 1 to 3. In Stage 3, Car 2 abruptly starts passing through the conflict area. Although the timing-taker increasingly favors the planner's conservative strategy earlier (mid-Stage 3), the planner does not immediately decelerate due to prediction bias and trajectory smoothing requirements. Full emergency braking occurs mid-Stage 4, ultimately causing a collision. In the second scenario as shown in Fig.\ref{fig:fig7}(b), during Stage 4, although the timing-taker recognizes that a cautious baseline strategy should be used, the planner fails to account for potential trajectory conflicts due the Car 1's nearly parallel to the ego vehicle, resulting in a collision with Car 1. These analyses reveal that the model's performance can be improved by using a more advanced base planner. To further identify the responsibility of different model components in collision events, we analyze the agent's average usage of planning-based actions over time, quantified by $\bar{\beta}^-=1-\bar{\beta}$, where $\bar{\beta}$ is the average value of $\beta$ calculated by Eq.\ref{eq:eq11}. In the test cases for the intersection scenario, the average $\bar{\beta}^-$ is $49.5\%$ across all timesteps, increasing to $65.7\%$ within $3$ seconds before a collision and reaching $85.9\%$ one second before the collision. In roundabout scenario, the corresponding values are $66.3\%$, $83.9\%$, and $87.2\%$, respectively. These results suggest that timing-aware RL can proactively adopt conservative baseline strategy to adapt to risky situations. Incorporting other robust and safety-focused methods as the baseline strategy instead of simple lattice planner, could further enhance the performace of the proposed method in the real-world autonomous driving application.

\section{Conclusion}

In this study, we propose a timing-aware reinforcement learning method, involving an "actor" to decide the suitable action at each moment, and a "timing-taker" to evaluate the optimal execution time of the action in the "timing imagination", which then provides safety constraints for the actor's action. A planning-based method is used as a safe baseline strategy for reacting cautiously to the dynamic environment when it is not the optimal time to execute the learned action. The timing-aware RL sufficiently combines the robustness of planning-based methods with the self-evolutionary capabilities of learning-based methods, and the alternating optimization in both reward and timing dimensions endows the model with a strong adaptability to highly dynamic environments.

The validation experiments choose the most typical driving scenarios with frequent inter-vehicle interactions -- an unsignalized intersection and a roundabout. In these two scenarios, the proposed model demonstrates better safety performance than the advanced safe RL models and effectively addresses the "action freezing" problem of the planning-based method in complex traffic environments. By observing the agent's action selection during the whole driving process, it can be found that the proposed method can perfectly balance the over-conservative behavior under the rule constraint and the over-aggressive motivated by short-term rewards, and emerges the micro-actions and social cues simliar to those that might occur in human's gaming process.

Limited by the simple implementation of the baseline model, there is still room to improve the task success rate of the proposed model in interaction scenarios. By analyzing the failure cases, we find that the errors of the baseline model account for a large proportion of the blame for the collisions, and the performance of the model can be significantly improved by adopting more powerful planning methods or end-to-end models in the future. In addition, the model implementations of both "actor" and "timing-taker" use only the simple MLP-based SAC architecture, and driving safety can be also improved by integrating the mechanism of timing-aware RL with other safe RL methods and applying more powerful neural network architecture (e.g., diffusion model). Last but not least, the method can also be applied to a wide range of motion planning and robotic tasks.



\bibliographystyle{unsrt}  
\bibliography{references}

\begin{thebibliography}{10}

\bibitem{feng2023dense}
Shuo Feng, Haowei Sun, Xintao Yan, Haojie Zhu, Zhengxia Zou, Shengyin Shen, and
  Henry~X Liu.
\newblock Dense reinforcement learning for safety validation of autonomous
  vehicles.
\newblock {\em Nature}, 615(7953):620--627, 2023.

\bibitem{hu2023planning}
Yihan Hu, Jiazhi Yang, Li~Chen, Keyu Li, Chonghao Sima, Xizhou Zhu, Siqi Chai,
  Senyao Du, Tianwei Lin, Wenhai Wang, et~al.
\newblock Planning-oriented autonomous driving.
\newblock In {\em Proceedings of the IEEE/CVF Conference on Computer Vision and
  Pattern Recognition}, pages 17853--17862, 2023.

\bibitem{huang2023differentiable}
Zhiyu Huang, Haochen Liu, Jingda Wu, and Chen Lv.
\newblock Differentiable integrated motion prediction and planning with
  learnable cost function for autonomous driving.
\newblock {\em IEEE transactions on neural networks and learning systems},
  2023.

\bibitem{silver2017mastering}
David Silver, Julian Schrittwieser, Karen Simonyan, Ioannis Antonoglou, Aja
  Huang, Arthur Guez, Thomas Hubert, Lucas Baker, Matthew Lai, Adrian Bolton,
  et~al.
\newblock Mastering the game of go without human knowledge.
\newblock {\em nature}, 550(7676):354--359, 2017.

\bibitem{kaufmann2023champion}
Elia Kaufmann, Leonard Bauersfeld, Antonio Loquercio, Matthias M{\"u}ller,
  Vladlen Koltun, and Davide Scaramuzza.
\newblock Champion-level drone racing using deep reinforcement learning.
\newblock {\em Nature}, 620(7976):982--987, 2023.

\bibitem{cao2023continuous}
Zhong Cao, Kun Jiang, Weitao Zhou, Shaobing Xu, Huei Peng, and Diange Yang.
\newblock Continuous improvement of self-driving cars using dynamic
  confidence-aware reinforcement learning.
\newblock {\em Nature Machine Intelligence}, 5(2):145--158, 2023.

\bibitem{lu2023imitation}
Yiren Lu, Justin Fu, George Tucker, Xinlei Pan, Eli Bronstein, Rebecca Roelofs,
  Benjamin Sapp, Brandyn White, Aleksandra Faust, Shimon Whiteson, et~al.
\newblock Imitation is not enough: Robustifying imitation with reinforcement
  learning for challenging driving scenarios.
\newblock In {\em 2023 IEEE/RSJ International Conference on Intelligent Robots
  and Systems (IROS)}, pages 7553--7560. IEEE, 2023.

\bibitem{achiam2017constrained}
Joshua Achiam, David Held, Aviv Tamar, and Pieter Abbeel.
\newblock Constrained policy optimization.
\newblock In {\em International conference on machine learning}, pages 22--31.
  PMLR, 2017.

\bibitem{cao2021confidence}
Zhong Cao, Shaobing Xu, Huei Peng, Diange Yang, and Robert Zidek.
\newblock Confidence-aware reinforcement learning for self-driving cars.
\newblock {\em IEEE Transactions on Intelligent Transportation Systems},
  23(7):7419--7430, 2021.

\bibitem{yang2021wcsac}
Qisong Yang, Thiago~D Sim{\~a}o, Simon~H Tindemans, and Matthijs~TJ Spaan.
\newblock Wcsac: Worst-case soft actor critic for safety-constrained
  reinforcement learning.
\newblock In {\em Proceedings of the AAAI Conference on Artificial
  Intelligence}, volume~35, pages 10639--10646, 2021.

\bibitem{ha2018recurrent}
David Ha and J{\"u}rgen Schmidhuber.
\newblock Recurrent world models facilitate policy evolution.
\newblock {\em Advances in neural information processing systems}, 31, 2018.

\bibitem{wang2022dynamic}
Junjie Wang, Qichao Zhang, and Dongbin Zhao.
\newblock Dynamic-horizon model-based value estimation with latent imagination.
\newblock {\em IEEE Transactions on Neural Networks and Learning Systems},
  2022.

\bibitem{wu2023daydreamer}
Philipp Wu, Alejandro Escontrela, Danijar Hafner, Pieter Abbeel, and Ken
  Goldberg.
\newblock Daydreamer: World models for physical robot learning.
\newblock In {\em Conference on Robot Learning}, pages 2226--2240. PMLR, 2023.

\bibitem{desjardins2011cooperative}
Charles Desjardins and Brahim Chaib-Draa.
\newblock Cooperative adaptive cruise control: A reinforcement learning
  approach.
\newblock {\em IEEE Transactions on intelligent transportation systems},
  12(4):1248--1260, 2011.

\bibitem{mirchevska2018high}
Branka Mirchevska, Christian Pek, Moritz Werling, Matthias Althoff, and Joschka
  Boedecker.
\newblock High-level decision making for safe and reasonable autonomous lane
  changing using reinforcement learning.
\newblock In {\em 2018 21st International Conference on Intelligent
  Transportation Systems (ITSC)}, pages 2156--2162. IEEE, 2018.

\bibitem{bautista2022autonomous}
Rolando Bautista-Montesano, Renato Galluzzi, Kangrui Ruan, Yongjie Fu, and Xuan
  Di.
\newblock Autonomous navigation at unsignalized intersections: A coupled
  reinforcement learning and model predictive control approach.
\newblock {\em Transportation research part C: emerging technologies},
  139:103662, 2022.

\bibitem{li2024nash}
Lin Li, Wanzhong Zhao, Chunyan Wang, Abbas Fotouhi, and Xuze Liu.
\newblock Nash double q-based multi-agent deep reinforcement learning for
  interactive merging strategy in mixed traffic.
\newblock {\em Expert Systems with Applications}, 237:121458, 2024.

\bibitem{yuan2021deep}
Mingfeng Yuan, Jinjun Shan, and Kevin Mi.
\newblock Deep reinforcement learning based game-theoretic decision-making for
  autonomous vehicles.
\newblock {\em IEEE Robotics and Automation Letters}, 7(2):818--825, 2021.

\bibitem{rahmati2021helping}
Yalda Rahmati, Mohammadreza~Khajeh Hosseini, and Alireza Talebpour.
\newblock Helping automated vehicles with left-turn maneuvers: A game
  theory-based decision framework for conflicting maneuvers at intersections.
\newblock {\em IEEE transactions on intelligent transportation systems},
  23(8):11877--11890, 2021.

\bibitem{wang2022comprehensive}
Xinpeng Wang, Songan Zhang, and Huei Peng.
\newblock Comprehensive safety evaluation of highly automated vehicles at the
  roundabout scenario.
\newblock {\em IEEE Transactions on Intelligent Transportation Systems},
  23(11):20873--20888, 2022.

\bibitem{chen2019attention}
Yilun Chen, Chiyu Dong, Praveen Palanisamy, Priyantha Mudalige, Katharina
  Muelling, and John~M Dolan.
\newblock Attention-based hierarchical deep reinforcement learning for lane
  change behaviors in autonomous driving.
\newblock In {\em Proceedings of the IEEE/CVF Conference on Computer Vision and
  Pattern Recognition Workshops}, pages 0--0, 2019.

\bibitem{cai2022dq}
Peide Cai, Hengli Wang, Yuxiang Sun, and Ming Liu.
\newblock Dq-gat: Towards safe and efficient autonomous driving with deep
  q-learning and graph attention networks.
\newblock {\em IEEE Transactions on Intelligent Transportation Systems},
  23(11):21102--21112, 2022.

\bibitem{huang2022efficient}
Zhiyu Huang, Jingda Wu, and Chen Lv.
\newblock Efficient deep reinforcement learning with imitative expert priors
  for autonomous driving.
\newblock {\em IEEE Transactions on Neural Networks and Learning Systems},
  2022.

\bibitem{kulkarni2016hierarchical}
Tejas~D Kulkarni, Karthik Narasimhan, Ardavan Saeedi, and Josh Tenenbaum.
\newblock Hierarchical deep reinforcement learning: Integrating temporal
  abstraction and intrinsic motivation.
\newblock {\em Advances in neural information processing systems}, 29, 2016.

\bibitem{le2018hierarchical}
Hoang Le, Nan Jiang, Alekh Agarwal, Miroslav Dud{\'\i}k, Yisong Yue, and Hal
  Daum{\'e}~III.
\newblock Hierarchical imitation and reinforcement learning.
\newblock In {\em International conference on machine learning}, pages
  2917--2926. PMLR, 2018.

\bibitem{cao2020reinforcement}
Zhangjie Cao, Erdem B{\i}y{\i}k, Woodrow~Z Wang, Allan Raventos, Adrien Gaidon,
  Guy Rosman, and Dorsa Sadigh.
\newblock Reinforcement learning based control of imitative policies for
  near-accident driving.
\newblock {\em arXiv preprint arXiv:2007.00178}, 2020.

\bibitem{oh2018self}
Junhyuk Oh, Yijie Guo, Satinder Singh, and Honglak Lee.
\newblock Self-imitation learning.
\newblock In {\em International conference on machine learning}, pages
  3878--3887. PMLR, 2018.

\bibitem{naveed2021trajectory}
Kaleb~Ben Naveed, Zhiqian Qiao, and John~M Dolan.
\newblock Trajectory planning for autonomous vehicles using hierarchical
  reinforcement learning.
\newblock In {\em 2021 IEEE International Intelligent Transportation Systems
  Conference (ITSC)}, pages 601--606. IEEE, 2021.

\bibitem{chen2019model}
Jianyu Chen, Bodi Yuan, and Masayoshi Tomizuka.
\newblock Model-free deep reinforcement learning for urban autonomous driving.
\newblock In {\em 2019 IEEE intelligent transportation systems conference
  (ITSC)}, pages 2765--2771. IEEE, 2019.

\bibitem{ma2021reinforcement}
Xiaobai Ma, Jiachen Li, Mykel~J Kochenderfer, David Isele, and Kikuo Fujimura.
\newblock Reinforcement learning for autonomous driving with latent state
  inference and spatial-temporal relationships.
\newblock In {\em 2021 IEEE International Conference on Robotics and Automation
  (ICRA)}, pages 6064--6071. IEEE, 2021.

\bibitem{chen2021interpretable}
Jianyu Chen, Shengbo~Eben Li, and Masayoshi Tomizuka.
\newblock Interpretable end-to-end urban autonomous driving with latent deep
  reinforcement learning.
\newblock {\em IEEE Transactions on Intelligent Transportation Systems},
  23(6):5068--5078, 2021.

\bibitem{ma2021model}
Haitong Ma, Jianyu Chen, Shengbo Eben, Ziyu Lin, Yang Guan, Yangang Ren, and
  Sifa Zheng.
\newblock Model-based constrained reinforcement learning using generalized
  control barrier function.
\newblock In {\em 2021 IEEE/RSJ International Conference on Intelligent Robots
  and Systems (IROS)}, pages 4552--4559. IEEE, 2021.

\bibitem{cheng2019end}
Richard Cheng, G{\'a}bor Orosz, Richard~M Murray, and Joel~W Burdick.
\newblock End-to-end safe reinforcement learning through barrier functions for
  safety-critical continuous control tasks.
\newblock In {\em Proceedings of the AAAI conference on artificial
  intelligence}, volume~33, pages 3387--3395, 2019.

\bibitem{li2021safe}
Jinning Li, Liting Sun, Jianyu Chen, Masayoshi Tomizuka, and Wei Zhan.
\newblock A safe hierarchical planning framework for complex driving scenarios
  based on reinforcement learning.
\newblock In {\em 2021 IEEE International Conference on Robotics and Automation
  (ICRA)}, pages 2660--2666. IEEE, 2021.

\bibitem{zhang2023spatial}
Zhili Zhang, Songyang Han, Jiangwei Wang, and Fei Miao.
\newblock Spatial-temporal-aware safe multi-agent reinforcement learning of
  connected autonomous vehicles in challenging scenarios.
\newblock In {\em 2023 IEEE International Conference on Robotics and Automation
  (ICRA)}, pages 5574--5580. IEEE, 2023.

\bibitem{nguyen2023safe}
Hung~Duy Nguyen and Kyoungseok Han.
\newblock Safe reinforcement learning-based driving policy design for
  autonomous vehicles on highways.
\newblock {\em International Journal of Control, Automation and Systems},
  21(12):4098--4110, 2023.

\bibitem{li2022lane}
Guofa Li, Yifan Qiu, Yifan Yang, Zhenning Li, Shen Li, Wenbo Chu, Paul Green,
  and Shengbo~Eben Li.
\newblock Lane change strategies for autonomous vehicles: a deep reinforcement
  learning approach based on transformer.
\newblock {\em IEEE Transactions on Intelligent Vehicles}, 2022.

\bibitem{schmidt2022learn}
Lukas~M Schmidt, Sebastian Rietsch, Axel Plinge, Bjoern~M Eskofier, and
  Christopher Mutschler.
\newblock How to learn from risk: Explicit risk-utility reinforcement learning
  for efficient and safe driving strategies.
\newblock In {\em 2022 IEEE 25th International Conference on Intelligent
  Transportation Systems (ITSC)}, pages 1913--1920. IEEE, 2022.

\bibitem{peng2022model}
Baiyu Peng, Jingliang Duan, Jianyu Chen, Shengbo~Eben Li, Genjin Xie, Congsheng
  Zhang, Yang Guan, Yao Mu, and Enxin Sun.
\newblock Model-based chance-constrained reinforcement learning via separated
  proportional-integral lagrangian.
\newblock {\em IEEE Transactions on Neural Networks and Learning Systems},
  2022.

\bibitem{gu2023safe}
Ziqing Gu, Lingping Gao, Haitong Ma, Shengbo~Eben Li, Sifa Zheng, Wei Jing, and
  Junbo Chen.
\newblock Safe-state enhancement method for autonomous driving via direct
  hierarchical reinforcement learning.
\newblock {\em IEEE Transactions on Intelligent Transportation Systems}, 2023.

\bibitem{wang2023autonomous}
Xuesong Wang, Jiazhi Zhang, Diyuan Hou, and Yuhu Cheng.
\newblock Autonomous driving based on approximate safe action.
\newblock {\em IEEE Transactions on Intelligent Transportation Systems}, 2023.

\bibitem{he2023fear}
Xiangkun He, Jingda Wu, Zhiyu Huang, Zhongxu Hu, Jun Wang, Alberto
  Sangiovanni-Vincentelli, and Chen Lv.
\newblock Fear-neuro-inspired reinforcement learning for safe autonomous
  driving.
\newblock {\em IEEE transactions on pattern analysis and machine intelligence},
  2023.

\bibitem{he2023towards}
Xiangkun He and Chen Lv.
\newblock Towards safe autonomous driving: Decision making with
  observation-robust reinforcement learning.
\newblock {\em Automotive Innovation}, 6(4):509--520, 2023.

\bibitem{zhou2022dynamically}
Weitao Zhou, Zhong Cao, Nanshan Deng, Xiaoyu Liu, Kun Jiang, and Diange Yang.
\newblock Dynamically conservative self-driving planner for long-tail cases.
\newblock {\em IEEE Transactions on Intelligent Transportation Systems},
  24(3):3476--3488, 2022.

\bibitem{cao2022trustworthy}
Zhong Cao, Shaobing Xu, Xinyu Jiao, Huei Peng, and Diange Yang.
\newblock Trustworthy safety improvement for autonomous driving using
  reinforcement learning.
\newblock {\em Transportation research part C: emerging technologies},
  138:103656, 2022.

\bibitem{haarnoja2018soft}
Tuomas Haarnoja, Aurick Zhou, Pieter Abbeel, and Sergey Levine.
\newblock Soft actor-critic: Off-policy maximum entropy deep reinforcement
  learning with a stochastic actor.
\newblock In {\em International conference on machine learning}, pages
  1861--1870. PMLR, 2018.

\bibitem{lopez2018microscopic}
Pablo~Alvarez Lopez, Michael Behrisch, Laura Bieker-Walz, Jakob Erdmann,
  Yun-Pang Fl{\"o}tter{\"o}d, Robert Hilbrich, Leonhard L{\"u}cken, Johannes
  Rummel, Peter Wagner, and Evamarie Wie{\ss}ner.
\newblock Microscopic traffic simulation using sumo.
\newblock In {\em 2018 21st international conference on intelligent
  transportation systems (ITSC)}, pages 2575--2582. IEEE, 2018.

\end{thebibliography}

\end{document}